\documentclass{article}

\usepackage{microtype}
\usepackage{graphicx}
\usepackage{subfigure}
\usepackage{booktabs} 

\usepackage{hyperref}

\usepackage[accepted]{icml2025}

\usepackage{amsmath}
\usepackage{amssymb}
\usepackage{mathtools}
\usepackage{amsthm}
\usepackage{enumitem}
\usepackage{listings}
\usepackage{booktabs}
\usepackage{multicol}
\usepackage{multirow}
\usepackage{graphicx}
\usepackage{tabularx}
\usepackage{arydshln}
\usepackage{float}
\usepackage{placeins}
\usepackage{xcolor}

\usepackage[capitalize,noabbrev]{cleveref}

\theoremstyle{plain}

\theoremstyle{definition}

\theoremstyle{remark}

\usepackage[textsize=tiny]{todonotes}

\icmltitlerunning{Compositional Translation: A Novel LLM-based Approach for Low-resource Machine Translation}

\DeclareUnicodeCharacter{02BB}{'}
\begin{document}

\twocolumn[
\icmltitle{Compositional Translation: A Novel LLM-based Approach\\ for Low-resource Machine Translation}



\icmlsetsymbol{equal}{*}

\begin{icmlauthorlist}
\icmlauthor{Armel Zebaze}{aaa}
\icmlauthor{Benoît Sagot}{aaa}
\icmlauthor{Rachel Bawden}{aaa}
\end{icmlauthorlist}

\icmlaffiliation{aaa}{Inria, Paris, France}

\icmlcorrespondingauthor{Armel Zebaze}{armel.zebaze-dongmo@inria.fr}

\icmlkeywords{Machine Learning, ICML}

\vskip 0.3in
]



\printAffiliationsAndNotice{}  

\begin{abstract}
The ability of generative large language models (LLMs) to perform in-context learning has given rise to a large body of research into how best to prompt models for various natural language processing tasks.
Machine Translation (MT) has been shown to benefit from in-context examples, in particular when they are semantically similar to the sentence to translate. In this paper, we propose a new LLM-based translation paradigm, \textit{compositional translation}, to replace naive few-shot MT with similarity-based demonstrations. An LLM is used to decompose a sentence into simpler phrases, and then to translate each phrase with the help of retrieved demonstrations. Finally, the LLM is prompted to translate the initial sentence with the help of the self-generated phrase-translation pairs. 
%
%
%
Our intuition is that this approach should improve translation because these shorter phrases should be intrinsically easier to translate and easier to match with relevant examples. This is especially beneficial in low-resource scenarios, and more generally whenever the selection pool is small or out of domain.
We show that \textit{compositional translation} boosts LLM translation performance on a wide range of popular MT benchmarks, including FLORES~200, NTREX~128 and TICO-19.
%
Code and outputs are available at \url{https://github.com/ArmelRandy/compositional-translation}.
\end{abstract}

\section{Introduction}

\begin{figure}[ht!!]
\begin{center}
    \centerline{\includegraphics[width=\linewidth]{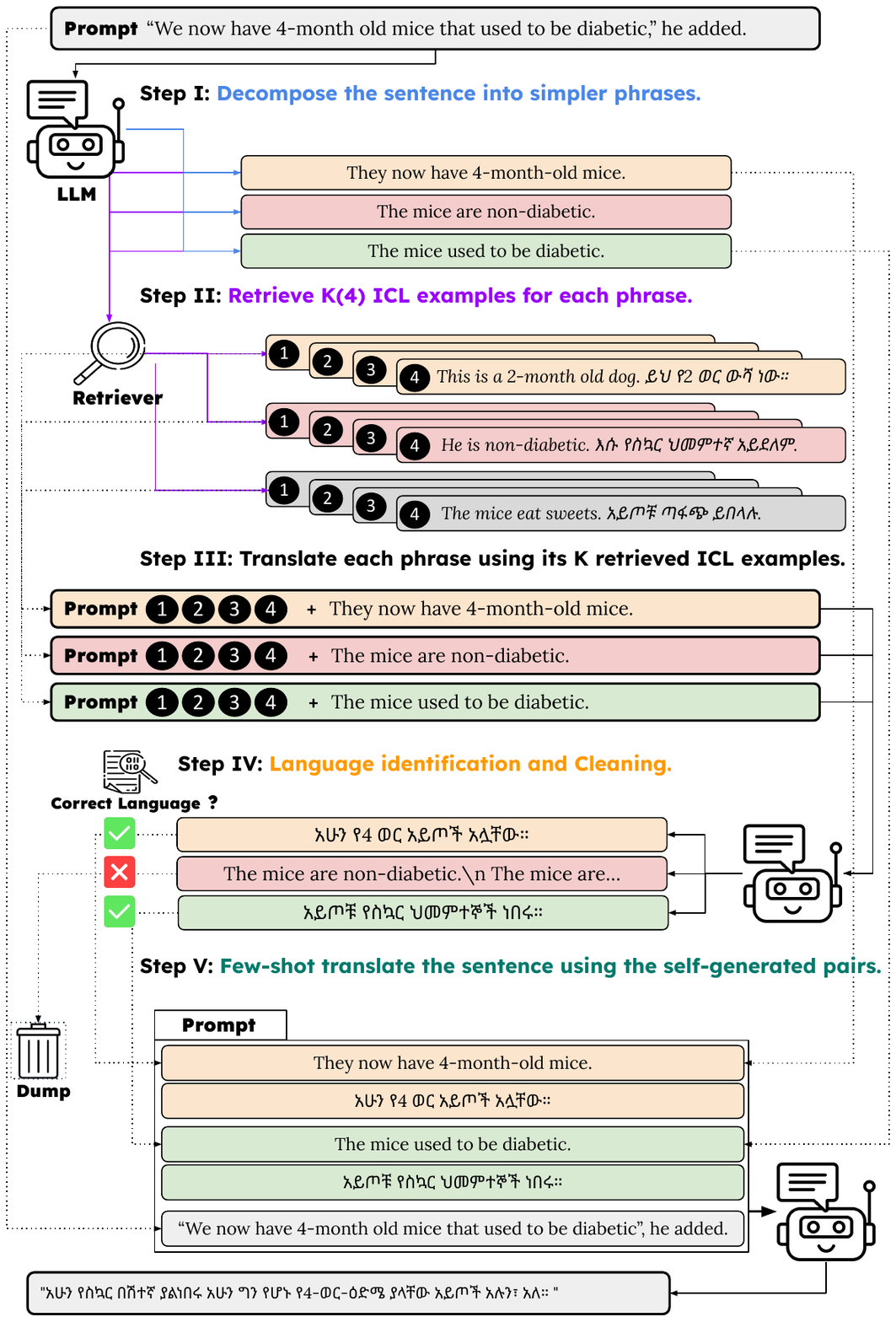}}
    \caption{An overview of \textbf{Compositional Translation (CompTra)}. Given a sentence, our method prompts the LLM to decompose it into several phrases that use its words. For each phrase, we retrieve relevant in-context demonstrations (here four) through similarity search and use them to translate it in a few-shot setup. The phrase-translation couples obtained are then cleaned and provided to the LLM to help it translate the main sentence.}
    \label{fig:fig1}
\end{center}
\vskip -0.2in
\end{figure}

Large Language Models (LLMs) have demonstrated strong performance across a wide variety of tasks~\citep{JMLR:v24:22-1144, dubey2024llama3herdmodels}. They use In-Context Learning (ICL; \citealp{NEURIPS2020_1457c0d6}) to solve problems at inference with the help of a few examples within their context. Multiple strategies have been introduced to expand the range of complexity of problems that can be addressed using ICL, with a particular emphasis on reasoning tasks~\citep{NEURIPS2022_9d560961, NEURIPS2022_8bb0d291, NEURIPS2023_271db992}. In machine translation (MT), most works have focused on choosing the best in-context examples. The prevailing intuition is to choose them based on their word-level similarity to the sentence to translate~\citep{agrawal-etal-2023-context, 10.5555/3618408.3620130, moslem-etal-2023-adaptive, DBLP:conf/acl/VilarFCLRF23, zhu-etal-2024-multilingual, bouthors-etal-2024-retrieving}. This intuition implicitly relies on the property of compositionality of MT~\citep{turcato-popowich-2001-example}; the translation of a sequence of words can be modeled as a function of the translation of its subparts. Choosing in-context examples based on similarity search amounts to finding subparts of a sentence to translate in another sentence that has a known translation.

While most studies only marginally explore LLMs for MT into low-resource languages (LRLs), we place them at the center of this work for two reasons: (i)~although current LLMs have narrowed the gap with supervised MT models for high-resource language (HRL) directions, they still struggle when translating into LRLs~\citep{gpt-mt-2023, enis2024llmnmtadvancinglowresource}, and (ii)~a few works have shown that example selection based on similarity can improve translation into LRLs~\citep{moslem-etal-2023-adaptive, tanzer2024a, zebaze2024incontextexampleselectionsimilarity}, suggesting a promising avenue for further research. We propose \textbf{Compositional Translation (CompTra)}, an LLM-based MT paradigm where translation is treated as a complex step by step reasoning task via the decomposition of the sentence into shorter and simpler entities. \citet{puduppully-etal-2023-decomt} proposed DecoMT as a decomposition-based approach to MT, in which a sentence is divided into subparts at the token-level and the translation is obtained by progressively solving fill-in-the-middle tasks. However their decomposition is uninformed, yielding incongruous subparts hard to translate and the sequential nature of their approach makes it slow and suboptimal for decoder-based models. Similarly, \citet{ghazvininejad2023dictionarybasedphraselevelpromptinglarge} and \citet{lu-etal-2024-chain} proposed to extract keywords from a sentence and translate them respectively with a dictionary and a multilingual dictionary. While the pieces obtained after decomposition are sound, their size do not provide enough context for accurate translation using an LLM. Moreover, their reliance on a dictionary considerably limits the impact of the intrinsic capabilities of the LLM on the MT task. In CompTra, given a sentence to translate, we prompt an LLM to generate simpler, concise and independent phrases that capture some of its aspects and use its words in the same context. We then proceed to self-generate their translations in few-shot, with demonstrations retrieved via similarity-search in a selection pool (tightening the constraint on the access to outside knowledge). Finally, the LLM derives the final translation by drawing insights from the above self-generated translation pairs. The underlying idea is that (i)~LLMs are more effective at handling short phrases, and (ii)~it is easier to retrieve relevant in-context demonstrations for translating such segments. These phrase translations will be of higher quality, and the similarity between the phrases and the source sentence will enable LLMs to produce better translations. 
Moreover, when the selection pool is small and lacks diversity, these self-generated phrase-translation pairs ensure high-quality and similar in-context demonstrations for the main translation. This contrasts with example selection via similarity search, which may not always provide the same level of similarity.

We evaluate CompTra on MT from English to LRLs: 10 languages from FLORES~200~\citep{goyal-etal-2022-flores, nllb2022} and 10 languages from NTREX~128~\citep{federmann-etal-2022-ntrex} and TICO-19~\citep{anastasopoulos-etal-2020-tico}. Our experiments with \texttt{Command-R+}, \texttt{LLaMA 3.1 70B It} and \texttt{Gemma 2 27B It} show that CompTra consistently outperforms example selection via similarity search and many existing strategies.


\section{Related Work}
\label{section:related_work}

\paragraph{Using LLMs for Machine Translation}
Similar to natural language understanding, reasoning and code generation, LLMs have been successfully applied to MT. \citet{NEURIPS2020_1457c0d6} showed that GPT-3 few-shot performance was on par with the state of the art for some translation directions at the time of its release. \citet{bawden-yvon-2023-investigating}, \citet{DBLP:conf/acl/VilarFCLRF23} and \citet{10.5555/3618408.3620130} explored aspects of few-shot MT including template selection and in-context example selection with BLOOM~\citep{workshop2023bloom}, PaLM~\citep{JMLR:v24:22-1144} and GLM~130B~\citep{zeng2023glmb} respectively. \citet{agrawal-etal-2023-context, mu-etal-2023-augmenting, bouthors-etal-2024-retrieving} documented performance gains when choosing in-context demonstrations semantically related to the sentence to translate. \citet{gpt-mt-2023} demonstrated that GPT models perform well as zero- and few-shot translators but face challenges with LRL pairs. Building on this, \citet{zhu-etal-2024-multilingual} conducted an extensive analysis across 102 languages, confirming these findings. They also noted that similarity search in a high-quality candidate pool offers no significant benefit. In contrast, \citet{zebaze2024incontextexampleselectionsimilarity} highlighted that while similarity-based selection does not help HRLs, it significantly improves performance for LRLs.

\vspace{-5pt}
\paragraph{Prompting and Compositionality}
After \citet{NEURIPS2020_1457c0d6} demonstrated the ability of LLMs to use ICL on diverse tasks, the research community investigated the development of better performing strategies for problem-solving at inference time. \citet{NEURIPS2022_9d560961} introduced chain-of-thought (CoT) prompting, which helps LLMs to mimic a step-by-step thought process by providing reasoning steps in the demonstrations. Following this, multiple works emerged on the necessity of the in-context examples for CoT prompting~\citep{NEURIPS2022_8bb0d291} and on their design~\citep{zhang2023automatic, fu2023complexitybased, yasunaga2024large}. More advanced techniques include self-consistency~\citep{wang2023selfconsistency} and hierarchical approaches such as Tree of Thoughts (ToT) \citep{NEURIPS2023_271db992} and Graph of Thoughts (GoT) \citep{Besta_Blach_Kubicek_Gerstenberger_Podstawski_Gianinazzi_Gajda_Lehmann_Niewiadomski_Nyczyk_Hoefler_2024}. Another line of research involves teaching LLMs to tackle complex problems by breaking them down into a series of subproblems and recursively solving them to derive the final answer~\citep{dua-etal-2022-successive, zhou2023leasttomost, khot2023decomposed, zebaze-etal-2024-tree}. All these efforts consistently enhanced the reasoning abilities of LLMs but had a limited effect on the MT task.

\vspace{-5pt}
\paragraph{Prompting LLMs for Machine Translation}
Beyond example selection for few-shot MT, several works have proposed prompting strategies for MT. \citet{puduppully-etal-2023-decomt} proposed DecoMT, which decomposes a sentence to translate into chunks of tokens, independently translate them, and derives the final translation by contextually translating each chunk one after another. The contextual translation of a chunk is obtained by using the contextual translation of the previous chunk as the left context and the independent translation of the next chunk as the right context. This inherently limits DecoMT's applicability to models trained like T5~\citep{raffel2020exploring} or trained with the Fill-In-the-Middle (FIM) objective~\citep{bavarian2022efficient}. While our work shares the idea of decomposition, we seek to derive simple, well-formed and coherent phrases that can be accurately translated independently from each other and directly used in few-shot for the main translation, making our approach non-sequential and thus faster. We propose a decomposition into subparts depending on the structure of the sentence (thus hyperparameter-free) where words in common with the main sentence are used in the same context with the end-goal of leveraging the property of compositionality of MT.
\citet{ghazvininejad2023dictionarybasedphraselevelpromptinglarge} introduced Dictionary-based Prompting for MT (DiPMT), which uses a dictionary to provide the target translations of certain words within a source sentence. These translations are incorporated into the input to help the LLM generate better translations. Building on this idea, \citet{lu-etal-2024-chain} proposed Chain-of-Dictionary (CoD), which extends DiPMT by translating chunks of words into multiple auxiliary languages and the target language using a multilingual dictionary, such as NLLB~\citep{nllb2022}. These translations are provided as additional context to further improve translation quality.
In contrast, our target is to improve LLM-based translation quality by only relying on the LLM itself.
Another line of research involves progressively guiding LLMs to produce good translations through a self-refinement process with or without external feedback \citep{chen2024iterativetranslationrefinementlarge, feng2024tearimprovingllmbasedmachine, xu-etal-2024-llmrefine, ki-carpuat-2024-guiding} inspired by the success of this paradigm for reasoning tasks~\citep{NEURIPS2023_91edff07, 10.5555/3666122.3666499}. This refining step is included in many strategies for MT.  \citet{briakou-EtAl:2024:WMT} proposed ``Translating Step-by-Step'' (SBYS): a multi-turn interaction with an LLM 
that breaks down the translation process into four distinct stages: identification of challenging components, drafting, refinement and proofreading. \citet{feng2024tearimprovingllmbasedmachine} designed another multi-step approach, ``Translate, Estimate, and Refine'' (TEaR), where a model generates a draft translation, self-derives the MQM annotations of the draft with the help of few-shot examples and subsequently refines the translation based on these annotations. On a different note, \citet{he2024exploring} proposed ``Multi-Aspect Prompting and Selection'' (MAPS), an ensembling technique that involves prompting a LLM to analyze a sentence for translation by building knowledge across three key aspects: keywords~\citep{aycock-bawden-2024-topic}, topics, and relevant demonstrations. Each aspect guides the LLM in generating a candidate translation. The final translation is then selected from these three candidates, along with the zero-shot output, based on the highest COMET~QE~\citep{rei-etal-2020-unbabels} score relative to the source sentence.

\vspace{-5pt}
\paragraph{LLMs and Low-Resource Languages}
LLMs are trained on increasingly larger datasets in accordance with scaling laws~\citep{kaplan2020scalinglawsneurallanguage, hoffmann2022trainingcomputeoptimallargelanguage, dubey2024llama3herdmodels}, which has led to them becoming more multilingual~\citep{workshop2023bloom, cahyawijaya-etal-2024-llms, enis2024llmnmtadvancinglowresource}, whether intentionally or not.~\citet{JMLR:v22:20-1307} and \citet{nllb2022} developed supervised multilingual MT models that significantly improved the state-of-the-art for various LRLs. Building on these advancements, subsequent progress included the release of massively multilingual datasets~\citep{schwenk-etal-2021-ccmatrix, abadji-etal-2022-towards, imanigooghari-etal-2023-glot500, singh-etal-2024-aya, futeral2024moscarlargescalemultilingualmultimodal}, as well as efforts in multilingual and multitask fine-tuning~\citep{muennighoff-etal-2023-crosslingual, ayamodelinstructionfinetuned, lai-etal-2024-llms} and continual pre-training~\citep{xu2024a, xu2024contrastive, xu2024xalmaplugplay, dou2024sailoropenlanguagemodels}.

\section{Methodology}
\label{section:methodology}
We introduce \textbf{Compositional Translation (CompTra)}, an LLM-based translation paradigm that automatically allows LLMs to translate any sentence by reasoning on self-generated translation pairs tailored to its content. CompTra frames any translation problem as an explicit step-by-step procedure. It consists of three main stages.
%
\begin{enumerate}[noitemsep, topsep=0pt, leftmargin=*]
    \item \textbf{Decomposition.} Given that the use of 
    related
    in-context examples helps few-shot MT into LRLs, we want to derive pairs as closely related to the source sentence as possible. The aim of the decomposition is to create \textbf{simpler} phrases that share words with the source sentence. We achieve this with the help of a \texttt{divide prompt}, which contains examples that demonstrate the decomposition, followed by the source sentence. The examples are from the MinWikiSplit corpus~\citep{niklaus-etal-2019-minwikisplit}; a set of sentences broken down into minimal propositions.
    It is worth noting that the number of phrases obtained is not a hyperparameter; each sentence is decomposed into the number of phrases that fits its structure.
    
    \item \textbf{Translation.} The LLM independently translates each of the phrases obtained. The \texttt{translate prompt} uses some artifacts, typically few-shot examples chosen via similarity search with a retriever. In practice, the phrases' translations are often written in an incorrect target language. We filterout phrase translations in the incorrect language with the help of a language identifier.
    \item \textbf{Recombination.} The LLM is fed with the phrases obtained after \textbf{decomposition} and their self-generated translations combined into a \texttt{merge prompt}. In our experiments, this prompt has exactly the same structure as the \texttt{translate prompt} in order to decouple the gains seen from changes to the prompt.
\end{enumerate}

The only hyperparameter is the number of demonstrations per phrase $k$, which we set to 5 for all phrases.
The hypothesis is that LLMs translate short sentences more effectively than longer ones, especially in languages they marginally encountered during their training. CompTra's objective is to propagate this advantage of short entities to bigger ones via a three-step hierarchical approach.

\section{Experiments}
\label{section:experiments}

\subsection{Experimental setup}
\label{section:experimental_setup}

\paragraph{Datasets.} We work on MT from English (eng) to LRLs.
\begin{itemize}[noitemsep, topsep=0pt, leftmargin=*]
    \item \textbf{FLORES~200}~\citep{goyal-etal-2022-flores, nllb2022}.~This dataset consists of translations from web articles into 204 languages. These sentences are divided into two splits: dev and devtest. We use the FLORES~200 dev set (997 examples) as the selection pool and the FLORES~200 devtest set (1012 examples) for the evaluation.


    \item \textbf{NTREX~128}~\citep{federmann-etal-2022-ntrex, barrault-etal-2019-findings} is an MT benchmark derived from WMT19 news data translated by professional human translators. It contains 1997 parallel sentences and is recommended for the evaluation of from-English translation directions. We use the first 1000 sentence pairs for evaluation, and the last 997 sentence pairs as the selection pool.
    

    
    \item \textbf{TICO-19}~\citep{anastasopoulos-etal-2020-tico} is an MT benchmark comprising texts on the COVID-19 pandemic covering 35 languages. Its validation and test sets consist of 971 (used as a selection pool) and 2100 samples respectively.
\end{itemize}
\vspace{-7pt}
\paragraph{Models.}
We use \texttt{LLaMA 3.1 It}  (8B, 70B; \citealp{dubey2024llama3herdmodels}), \texttt{Gemma 2 It} (9B, 27B;  \citealp{gemmateam2024gemma2improvingopen}), 
 \texttt{Command-R} and \texttt{-R+}~\footnote{\texttt{command-r-08-2024}, \texttt{command-r-plus-08-2024}}\citep{Cohere_Command-R-plus}.
\vspace{-7pt}
\paragraph{Evaluation Metrics.}
We mainly evaluate using MetricX-23~\citep{juraska-etal-2023-metricx} and XCOMET~\citep{guerreiro-etal-2024-xcomet}, which are the highest ranked non-ensemble metrics according to the latest WMT shared task on MT metrics~\cite{freitag-EtAl:2024:WMT}. Precisely we use their reference-based versions \texttt{XCOMET-XXL} (which supports the same 100 languages as XLM~RoBERTa~\citep{conneau-etal-2020-unsupervised}) and \texttt{MetricX-23-XXL} (which supports the same 101 languages as mT5~\citep{xue2021mt5}). 
MetricX assigns a score ranging from 0 to 25, with higher scores indicating more errors in the translation. XCOMET produces a score between 0 and 1, which we rescale to a range of 0 to 100, where higher scores represent better translation quality.
We also consider $n$-gram matching metrics via sacreBLEU~\citep{post-2018-call}, namely chrF++\footnote{nrefs:1$|$case:mixed$|$eff:yes$|$nc:6$|$nw:2$|$space:no$|$version:2.4.2}~\citep{popovic-2015-chrf, popovic:2017:WMT} and BLEU\footnote{nrefs:1$|$case:mixed$|$eff:no$|$tok:flores200$|$smooth:exp$|$version:2.4.2}~\citep{10.3115/1073083.1073135}
for transparency reasons.
\vspace{-10pt}
\paragraph{Baselines.}
We compare to zero-shot and 5-shot MT with example retrieval via similarity with BM25~\citep{robertson1995okapi} and SONAR~\citep{duquenne2023sonarsentencelevelmultimodallanguageagnostic}. The former retrieves a candidate based on its BM25 score relative to the query, whereas the latter uses the cosine similarity between the query and the candidates in the embedding space.
\vspace{-10pt}
\paragraph{Experimental Details.}
In all experiments, CompTra uses BM25~\citep{robertson1995okapi} as its retriever and queries 5 in-context examples for each phrase unless specified otherwise. Language identification is done with FastText~\citep{bojanowski-etal-2017-enriching, nllb2022} and only when the language is supported. We use \texttt{vLLM} \citep{kwon2023efficient} for inference with greedy decoding and BM25s~\citep{bm25s}. We generate at most 500 new tokens during the translation phase and 2000 during combination. We remove repeating bigrams at the end of the translations.\footnote{See Appendix~\ref{appendix:general_remarks}.}

\subsection{Results} \label{sec:results}

\subsubsection{Main results: FLORES~200}
We evaluate CompTra on 10 English-to-X translation directions from FLORES~200  (Table~\ref{tab:flores200}).\footnote{See Appendix~\ref{appendix:bleu-chrf-flores} for BLEU and chrF++ scores.} CompTra consistently outperforms similarity-based few-shot MT across all directions and for all the LLMs we evaluated. On average, CompTra outperforms few-shot BM25 by 0.4 MetricX with \texttt{LLaMA 3.1 70B It} and \texttt{Gemma 2 27B It}, and by 1.5 MetricX with \texttt{Command-R+}. For XCOMET, the gains are 1.0 with \texttt{Gemma 2 27B It} and 1.8 with \texttt{Command-R+}.

\begin{table*}[pht]
\caption{Full quantitative results for 10 English $\xrightarrow{}$ X translation directions from FLORES~200~\citep{goyal-etal-2022-flores, nllb2022} (XCOMET and MetricX scores). Best results (including any results that are not statistically worse) are highlighted in bold.}
\label{tab:flores200}
\vskip 0.15in
\small
\begin{center}
\resizebox{\textwidth}{!}{
\begin{tabular}{lrrrrrrrrrrrrrr}
\toprule
\multirow{2}{*}{Methods}  & \multicolumn{2}{c}{Amharic} & & \multicolumn{2}{c}{Burmese} & & \multicolumn{2}{c}{Fijian} & & \multicolumn{2}{c}{Khmer} & & \multicolumn{2}{c}{Lao}\\
\cmidrule{2-3} \cmidrule{5-6} \cmidrule{8-9} \cmidrule{11-12} \cmidrule{14-15}
{} & {XCOMET} & {MetricX} &  & {XCOMET} & {MetricX} & & {XCOMET} & {MetricX} & & {XCOMET} & {MetricX} & & { XCOMET} & {MetricX}\\
\midrule
\multicolumn{15}{l}{LLaMA 3.1 70B Instruct} \\
\midrule
Zero-shot    & 31.25 & 16.95 &  & 57.73 & 5.18 & & 20.44 & 19.48 & & 60.76 & 5.69 & & 32.63 & 16.67 \\
5-shot SONAR & 37.51 & 13.61 &  & 60.91 & 4.50 & & 21.84 & 15.52 & & 64.00 & 4.92 & & 42.05 & 11.55 \\
5-shot BM25  & 39.55 & 13.02 &  & \bf 62.27 & 4.26 & & 22.18 & 15.23 & & \bf 64.46 & 4.92 & & 45.38 & 11.02 \\
CompTra (Ours) & \bf 41.32 & \bf 11.95 &  & 60.84 & \bf 3.64 & & \bf 22.39 & \bf 14.94 & & 64.22 & \bf 4.75 & & \bf 47.17 & \bf 10.54 \\
\midrule
\multicolumn{15}{l}{Gemma 2 27B It} \\
\midrule
Zero-shot    & 32.79 & 15.49 &  & 40.07 & 9.05 & & 19.91 & 20.39 & & 45.37 & 9.15 & & 38.80 & 13.40 \\
5-shot SONAR & 37.07 & 13.69 &  & 47.38 & 7.02 & & 21.00 & 18.20 & & 53.05 & 7.14 & & 47.17 & 10.58 \\
5-shot BM25  & 38.09 & 13.23 &  & \bf 48.62 & \bf 6.80 & & 20.96 & 18.17 & & 54.00 & \bf 7.02 & & 48.03 & 10.50 \\
CompTra (Ours) & \bf 40.10 & \bf 12.64 &  & 47.98 & 7.23 & & \bf 21.55 & \bf 16.86 & & \bf 55.02 & \bf 7.05 & & \bf 51.02 & \bf 9.88\\
\midrule
\multicolumn{15}{l}{Command-R+} \\
\midrule
Zero-shot    & 16.32 & 24.46 &  & 24.39 & 19.63 & & 18.63 & 22.95 & & 23.39 & 19.44 & & 19.81 & 21.71 \\
5-shot SONAR & 18.06 & 23.45 &  & 29.60 & 15.12 & & 20.13 & 19.52 & & 27.91 & 18.45 & & 25.48 & 18.46 \\
5-shot BM25  & 17.96 & 23.24 &  & 32.25 & 14.76 & & 20.30 & 19.19 & & 28.37 & 18.40 & & 26.74 & 18.03 \\
CompTra (Ours) & \bf 19.33 & \bf 22.54 &  & \bf 35.60 & \bf 12.34 & & \bf 21.16 & \bf 16.85 & & \bf 31.76 & \bf 15.53 & & \bf 29.64 & \bf 16.52\\

\end{tabular}
}
\resizebox{\textwidth}{!}{
\begin{tabular}{lrrrrrrrrrrrrrr}
\toprule
\multirow{2}{*}{Methods}  & \multicolumn{2}{c}{Samoan} & & \multicolumn{2}{c}{Sinhala} & & \multicolumn{2}{c}{Tsonga} & & \multicolumn{2}{c}{Turkmen} & & \multicolumn{2}{c}{Uyghur}\\
\cmidrule{2-3} \cmidrule{5-6} \cmidrule{8-9} \cmidrule{11-12} \cmidrule{14-15}
{} & {XCOMET} & {MetricX} &  & {XCOMET} & {MetricX} & & {XCOMET} & {MetricX} & & {XCOMET} & {MetricX} & & { XCOMET} & {MetricX}\\
\midrule
\multicolumn{15}{l}{LLaMA 3.1 70B Instruct} \\
\midrule
Zero-shot    & 23.61 & 11.51&  & 64.72 & 4.27 & & 21.91 & 19.82 & & 23.46 & 8.77 & & 43.65 & 8.36 \\
5-shot SONAR & 24.42 & 9.30 &  & 67.60 & 3.60 & & 23.66 & 16.45 & & 25.14 & 5.83 & & 58.16 & \bf 4.33 \\
5-shot BM25  & 24.78 & 9.14 &  & 67.64 & 3.64 & & 24.46 & 15.88 & & 24.97 & 5.66 & & \bf 59.70 & 4.37 \\
CompTra (Ours) & \bf 24.95 & \bf 8.89 &  & \bf 68.63 & \bf 3.28 & & \bf 24.77 & \bf 15.38 & & \bf 25.40 & \bf 5.14 & & 58.32 & \bf 4.30 \\
\midrule
\multicolumn{15}{l}{Gemma 2 27B It} \\
\midrule
Zero-shot    & 21.38 & 17.89 &  & 48.27 & 7.48 & & 21.81 & 20.02 & & 25.26 & 6.07 & & 29.83 & 12.90 \\
5-shot SONAR & 22.30 & 15.16 &  & 52.30 & 6.65 & & 23.84 & 16.39 & & 25.78 & 4.53 & & 36.34 & 9.98 \\
5-shot BM25  & 22.69 & 14.59 &  & 53.68 & 6.50 & & 24.06 & 15.97 & & \bf 26.26 & \bf 4.48 & & 37.52 & 9.72 \\
CompTra (Ours) & \bf 23.16 & \bf 13.80 &  & \bf 54.47 & \bf 6.19 & & \bf 24.53 & \bf 15.02 & & 25.98 & 4.55 & & \bf 38.97 & \bf 9.37\\
\midrule
\multicolumn{15}{l}{Command-R+} \\
\midrule
Zero-shot    & 18.84 & 23.21 &  & 38.56 & 10.62 & & 19.01 & 24.36 & & 25.03 & 6.04 & & 29.12 & 14.00 \\
5-shot SONAR & 20.30 & 20.15 &  & 45.60 & 8.65 & & 20.56 & 22.41 & & 24.93 & 4.70 & & 39.74 & 9.46 \\
5-shot BM25  & 20.75 & 19.48 &  & 47.06 & 8.11 & & 20.83 & 22.08 & & 25.11 & 4.57 & & 42.00 & 8.54 \\
CompTra (Ours) & \bf 21.78 & \bf 17.31 &  & \bf 49.29 & \bf 6.99 & & \bf 21.89 & \bf 21.59 & & \bf 25.26 & \bf 4.07 & & \bf 44.14 & \bf 7.64 \\
\bottomrule
\end{tabular}
}
\end{center}
\vskip -0.1in
\end{table*}

\subsubsection{Results on NTREX~128 and TICO-19}
We report the results obtained by CompTra on NTREX~128 and TICO-19 in Tables~\ref{tab:ntrex}\footnote{See Appendix~\ref{appendix:bleu-chrf-ntrex} for BLEU and chrF++ scores.} and~\ref{tab:tico} respectively. 
Similar to our observations on FLORES-200, CompTra consistently outperforms similarity-based few-shot MT across all directions and for all the LLMs.
\begin{table*}[pht]
\caption{Full XCOMET and MetricX results for 5 English$\xrightarrow{}$X directions from NTREX~128~\citep{federmann-etal-2022-ntrex}.}
\label{tab:ntrex}
\small
\begin{center}
\resizebox{\textwidth}{!}{
\begin{tabular}{lrrrrrrrrrrrrrr}
\toprule
\multirow{2}{*}{Methods}  & \multicolumn{2}{c}{Amharic} & & \multicolumn{2}{c}{Fijian} & & \multicolumn{2}{c}{Shona} & & \multicolumn{2}{c}{Somali} & & \multicolumn{2}{c}{Tswana}\\
\cmidrule{2-3} \cmidrule{5-6} \cmidrule{8-9} \cmidrule{11-12} \cmidrule{14-15}
{} & XCOMET  & MetricX &  & XCOMET & MetricX & & XCOMET & MetricX & & XCOMET & MetricX & & XCOMET & MetricX-23\\
\midrule
LLaMA~3.1~70B~It \\
\midrule
5-shot BM25 & 30.62 & 16.14 &  & 21.61 & 15.80 & & 24.43 & 15.52 & & 40.39 & 8.99 & & 26.24 & 12.40 \\
CompTra (Ours) & \bf 31.61 & \bf 15.46 &  & \bf 21.75 & \bf 15.38 & & \bf 24.69 & \bf 15.18 & & \bf 40.70 & \bf 8.85 & & \bf 26.73 & \bf 12.15 \\
\midrule
Gemma~2~27B~It \\
\midrule
5-shot BM25 & 29.67 & 16.01 &  & 20.54 & 18.34 & & 25.27 & 10.50 & & 41.89 & 8.25 & & 26.49 & 11.49 \\
CompTra (Ours) & \bf 30.46 & \bf 15.40 &  & \bf 21.04 & \bf 17.35 & & \bf 25.63 & 10.48 & & \bf 42.34 & \bf 8.11 & & \bf 26.69 & \bf 11.42 \\
\midrule
Command-R+ \\
\midrule
5-shot BM25 & 18.33 & 23.05 &  & 19.96 & 18.77 & & 21.42 & 19.26 & & 25.86 & 16.78 & & 22.55 & 18.94 \\
CompTra (Ours) & 17.80 & 22.98 &  & \bf 20.66 & \bf 17.42 & & \bf 22.21 & \bf 18.43 & & \bf 26.97 & \bf 15.53 & & \bf 23.95 & \bf 17.96\\
\bottomrule
\end{tabular}
}
\end{center}
\end{table*}

\begin{table*}[pht]
\caption{Full XCOMET and MetricX results for 5 English$\rightarrow$X translation directions from TICO-19 \citep{anastasopoulos-etal-2020-tico}.}
\small
\label{tab:tico}
\begin{center}
\resizebox{\textwidth}{!}{
\begin{tabular}{lrrrrrrrrrrrrrr}
\toprule
\multirow{2}{*}{Methods}  & \multicolumn{2}{c}{Amharic} & & \multicolumn{2}{c}{Khmer} & & \multicolumn{2}{c}{Lingala} & & \multicolumn{2}{c}{Luganda} & & \multicolumn{2}{c}{Tamil}\\
\cmidrule{2-3} \cmidrule{5-6} \cmidrule{8-9} \cmidrule{11-12} \cmidrule{14-15}
{} & XCOMET  & MetricX &  & XCOMET & MetricX & & XCOMET & MetricX & & XCOMET & MetricX & & XCOMET & MetricX-23\\
\midrule
LLaMA~3.1~70B~It \\
\midrule
5-shot BM25 & 39.71 & 12.71 &  & 67.55 & 4.51 & & \bf 23.71 & 14.69 & & 26.70 & 12.77 & & 68.00 & 2.06 \\
CompTra (Ours) & \bf 40.40 & \bf 11.83 &  & \bf 68.96 & \bf 3.72 & & 23.65 & \bf 14.53 & & 26.66 & \bf 12.68 & & \bf 68.46 & \bf 1.50 \\
\midrule
Gemma~2~27B~It \\
\midrule
5-shot BM25 & 40.80 & 11.99 &  & 60.42 & 5.71 & & 24.03 & \bf 13.90 & & 26.45 & 14.99 & & \bf 68.47 & \bf 1.50 \\
CompTra (Ours) & \bf 41.84 & \bf 11.30 &  & \bf 62.25 & \bf 5.38 & & \bf 24.21 & 13.98 & & \bf 26.67 & \bf 13.90 & & 67.41 & 1.62 \\
\midrule
Command-R+ \\
\midrule
5-shot BM25 & 22.45 & 21.40 &  & 38.65 & 13.69 & & \bf 22.68 & 16.94 & & 23.29 & 20.24 & & \bf 66.86 & 2.15 \\
CompTra (Ours) & \bf 23.03 & \bf 21.08 &  & \bf 40.17 & \bf 12.36 & & \bf 22.70 & \bf 16.22 & & \bf 23.93 & \bf 19.61 & & 66.34 & \bf 1.71 \\
\bottomrule
\end{tabular}
}
\end{center}
\vskip -0.1in
\end{table*}

\subsubsection{Comparison to existing approaches}
We conduct a set of additional studies and compare \textit{compositional translation} against existing methods including zero- and few-shot MT and CoT~\citep{NEURIPS2022_8bb0d291, peng-etal-2023-towards}, MAPS~\citep{he2024exploring}, TEaR~\cite{feng2024tearimprovingllmbasedmachine}, SBYS~\citep{briakou-EtAl:2024:WMT} and standalone Self-Refine~\citep{chen2024iterativetranslationrefinementlarge}. We use \texttt{LLaMA 3.1 70B It} and report the results in Table~\ref{tab:sota}. CompTra significantly outperforms all the strategies. 5-shot BM25 emerges as a very strong baseline which can be further improved via self-refine, although it does not reach CompTra's performance.

\begin{table*}[pht]
\caption{Full XCOMET and MetricX results 10 English$\rightarrow$X directions from FLORES~200. We compare CompTra to CoT~\citep{NEURIPS2022_8bb0d291}, MAPS~\citep{he2024exploring}, SBYS~\citep{briakou-EtAl:2024:WMT} and TEaR~\citep{feng2024tearimprovingllmbasedmachine}.}
\label{tab:sota}
\small
\begin{center}
\resizebox{\textwidth}{!}{
\begin{tabular}{lrrrrrrrrrrrrrr}
\toprule
\multirow{2}{*}{Methods}  & \multicolumn{2}{c}{Amharic} & & \multicolumn{2}{c}{Burmese} & & \multicolumn{2}{c}{Fijian} & & \multicolumn{2}{c}{Khmer} & & \multicolumn{2}{c}{Lao}\\
\cmidrule{2-3} \cmidrule{5-6} \cmidrule{8-9} \cmidrule{11-12} \cmidrule{14-15}
{} & XCOMET  & MetricX &  & XCOMET & MetricX & & XCOMET & MetricX & & XCOMET & MetricX & & XCOMET & MetricX-23\\
\midrule
Zero-shot           & 31.25 & 16.95 &  & 57.73 & 5.18 & & 20.44 & 19.48 & & 60.76 & 5.69 & & 32.63 & 16.67 \\
Zero-shot + CoT     & 23.35 & 22.62 &  & 38.61 &13.64 & & 19.97 & 20.55 & & 50.77 & 8.90 & & 22.98 & 22.83 \\
Zero-shot + Refine  & 32.21 & 16.56 &  & 58.92 & 4.76 & & 20.39 & 19.56 & & 61.39 & 5.28 & & 32.98 & 16.82 \\
SBYS                & 31.77 & 16.55 &  & 56.79 & 5.16 & & 20.28 & 19.27 & & 61.89 & 4.86 & & 31.58 & 16.40\\
MAPS                & 35.63 & 14.28 &  & 60.94 & 4.16 & & 20.37 & 19.15 & & 61.47 & 5.37 & & 37.69 & 14.41 \\
TEaR                & 37.96 & 13.69 &  & 59.89 & 6.02 & & 21.70 & 16.32 & & 63.49 & 5.13 & & 42.83 & 12.17 \\
\hdashline
5-shot BM25         & 39.55 & 13.03 &  & 62.27 & 4.26 & & 22.18 & 15.23 & & \bf 64.46 & 4.92 & & 45.38 & 11.02 \\
 + CoT              & 32.39 & 18.18 &  & 52.07 & 7.18 & & 22.08 & 15.97 & & 60.80 & 5.80 & & 36.78 & 15.90 \\
 + Refine           & 38.38 & 13.46 &  & \bf 62.97 & 3.82 & & 21.88 & 15.96 & & 64.39 & \bf 4.76 & & 43.94 & 11.59 \\
\hdashline
CompTra (Ours)        & \bf 41.32 & \bf 11.95 &  & 60.84 & \bf 3.64 & & \bf 22.39 & \bf 14.94 & & 64.22 & \bf 4.75 & & \bf 47.17 & \bf 10.54\\
\end{tabular}
}
\resizebox{\textwidth}{!}{
\begin{tabular}{lrrrrrrrrrrrrrr}
\toprule
\multirow{2}{*}{Methods}  & \multicolumn{2}{c}{Samoan} & & \multicolumn{2}{c}{Sinhala} & & \multicolumn{2}{c}{Tsonga} & & \multicolumn{2}{c}{Turkmen} & & \multicolumn{2}{c}{Uyghur}\\
\cmidrule{2-3} \cmidrule{5-6} \cmidrule{8-9} \cmidrule{11-12} \cmidrule{14-15}
{} & XCOMET  & MetricX &  & XCOMET & MetricX & & XCOMET & MetricX & & XCOMET & MetricX & & XCOMET & MetricX-23\\
\midrule
Zero-shot           & 23.61 & 11.51 &  & 64.72 & 4.27 & & 21.91 & 19.82 & & 23.46 & 8.77 & & 43.65 & 8.36 \\
Zero-shot + CoT     & 22.58 & 14.08 &  & 49.44 & 8.47 & & 21.33 & 21.45 & & 22.83 & 10.78 & & 38.25 & 11.53 \\
Zero-shot + Refine  & 23.49 & 11.26 &  & 66.27 & 4.11 & & 22.18 & 19.68 & & 23.65 & 7.77 & & 47.38 & 7.60 \\
SBYS                & 22.54 & 11.70 &  & 65.31 & 3.61 & & 21.13 & 20.25 & & 23.83 & 7.63 & & 47.82 & 6.28 \\
MAPS                & 23.12 & 11.89 &  & 68.14 & 3.41 & & 23.30 & 18.77 & & 24.44 & 9.03 & & 51.88 & 6.20 \\
TEaR                & 24.45 &  9.72 &  & 65.94 & 4.38 & & 24.01 & 16.34 & & 25.15 & 5.73 & & 57.20 & 5.06 \\
\hdashline
5-shot BM25         & 24.78 & 9.14 &  & 67.64 & 3.64 & & 24.46 & 15.88 & & 24.97 & 5.66 & & 59.70 & 4.37 \\
 + CoT              & 24.09 & 10.65 &  & 65.64 & 4.06 & & 24.08 & 17.22 & & 24.96 & 6.22 & & 52.46 & 6.10 \\
 + Refine           & 24.34 & 9.26 &  & \bf 69.13 & 3.49 & & 24.23 & 16.15 & & 24.74 & 5.26 & & \bf 60.50 & \bf 4.17 \\
\hdashline
CompTra (Ours)        & \bf 24.95 & \bf 8.89 &  & 68.63 & \bf 3.28 & & \bf 24.77 & \bf 15.38 & & \bf 25.40 & \bf 5.14 & & 58.32 & 4.30\\
\bottomrule
\end{tabular}
}
\end{center}
\vskip -0.1in
\end{table*}

\begin{table}[H]
\caption{Full MetricX results for ten English$\rightarrow$X directions with smaller LMs.}
\label{tab:scaling}
\small
\begin{center}
\resizebox{0.9\linewidth}{!}{
\begin{tabular}{lrrrrr}
\toprule
{} & Amharic & Burmese & Fijian & Khmer & Lao  \\
\midrule
\multicolumn{6}{l}{LLaMA 3.1 8B It} \\
\midrule
5-shot BM25 & 23.40 & \bf 14.27 & 21.74 & 12.63 & 22.81 \\
CompTra (Ours) & \bf 23.06 & \bf 14.29 & \bf 20.93 & \bf 12.02 & \bf 22.41  \\
\midrule
\multicolumn{6}{l}{Gemma 2 9B It} \\
\midrule
5-shot BM25 & 15.99 & 13.05 & 20.66 & 11.92 & 15.21  \\
CompTra (Ours) & \bf 15.66 & \bf 12.31 & \bf 19.63 & \bf 11.23 & \bf 13.67 \\
\midrule
\multicolumn{6}{l}{Command-R} \\
\midrule
5-shot BM25 & \bf 24.38 & 20.94 & 21.24 & 21.64 & 22.68 \\
CompTra (Ours) & \bf 24.39 & \bf 19.33 & \bf 20.59 & \bf 20.48 & \bf 21.88 \\
\end{tabular}
}
\resizebox{0.9\linewidth}{!}{
\begin{tabular}{lrrrrr}
\toprule
{} & Samoan & Sinhala & Tsonga & Turkmen & Uyghur \\
\midrule
\multicolumn{6}{l}{LLaMA 3.1 8B It} \\
\midrule
5-shot BM25 & 19.80 & 13.79 & 23.02 & 14.72 & \bf 14.01 \\
CompTra (Ours) & \bf 18.25 & \bf 13.23 & \bf 22.75 & \bf 14.39 & 15.00 \\
\midrule
\multicolumn{6}{l}{Gemma 2 9B It} \\
\midrule
5-shot BM25  & 17.61 & 9.13 &  20.99 & 8.36 & 21.07 \\
CompTra (Ours) & \bf 15.93 & \bf 8.82 & \bf 19.82 & \bf 7.69 & \bf 19.19 \\
\midrule
\multicolumn{6}{l}{Command-R} \\
\midrule
5-shot BM25  & 21.67 & 15.50 & 22.46 & 7.00 & 16.36 \\
CompTra (Ours) & \bf 20.82 & \bf 12.91 & \bf 22.16 & \bf 5.95 & \bf 14.99 \\
\bottomrule
\end{tabular}
}
\end{center}
\vskip -0.1in
\end{table}

\begin{table*}[ht]
\caption{Impact of switching few-shot MT with NLLB in CompTra (MetricX scores).}
\label{tab:nllb}
\tiny
\begin{center}
\resizebox{0.9\linewidth}{!}{
\begin{tabular}{lrrrrrrrrrr}
\toprule
{} & Amharic & Burmese & Fijian & Khmer & Lao & Samoan & Sinhala & Tsonga & Turkmen & Uyghur \\
\midrule
{NLLB-200-distilled-600M} & \bf 5.46 & 6.32 & \bf 9.69 & 9.92 & \textbf{5.75} & 6.25 & 5.16 & \bf 8.78 & 9.04 & 7.17 \\
\midrule
\multicolumn{11}{l}{LLaMA 3.1 70B Instruct} \\
\midrule
NLLB + CompTra & 5.58 & 5.06 & 10.12 & 6.72 & \textbf{5.75} & \bf 5.54 & \bf 3.19 & \bf 8.74 & 6.21 & 5.66 \\
5-shot BM25 & 13.02 & 4.26 & 15.23 & 4.92 & 11.02 & 9.14 & 3.64 & 15.88 & 5.66 & 4.37 \\
CompTra with BM25 & 11.95 & \bf 3.64 & 14.94 & \bf 4.75 & 10.54 & 8.89 & 3.28 & 15.38 & \bf 5.14 & \bf 4.30 \\
\bottomrule
\end{tabular}
}
\end{center}
\vskip -0.1in
\end{table*}

\begin{table*}[t!]
\caption{Ablation study on the impact of the retriever on CompTra (MetricX scores).}
\label{tab:retriever}
\tiny
\begin{center}
\resizebox{0.9\linewidth}{!}{
\begin{tabular}{lrrrrrrrrrr}
\toprule
{} & Amharic & Burmese & Fijian & Khmer & Lao & Samoan & Sinhala & Tsonga & Turkmen & Uyghur \\
\midrule
\multicolumn{11}{l}{LLaMA 3.1 70B Instruct} \\
\midrule
5-shot SONAR & 13.61 & 4.50 & \bf 15.52 & 4.92 & \bf 11.55 & \bf 9.30 & 3.60 & \bf16.44 & 5.83 & \bf 4.33 \\
CompTra with SONAR & \bf 12.93 & \bf 4.00 & 15.86 & 4.98 & 11.95 & 9.86 & \bf 3.27 & 16.86 & \bf 5.67 & 4.79 \\
\hdashline
5-shot LCS     & 14.74 & 6.42 & 16.89 & 5.40 & \bf 13.07 & \bf 10.14 & 3.97 & \bf 17.58 & 6.33 & \bf 4.79 \\
CompTra with LCS & \bf 14.65 & \bf 4.22 & \bf 16.74 & \bf 5.36 & 13.44 & 10.57 & \bf 3.59 & 17.82 & \bf 6.07 & 5.07 \\
\hdashline
5-shot BM25 & 13.02 & 4.26 & 15.23 & 4.92 & 11.02 & 9.14 & 3.64 & 15.88 & 5.66 & 4.37 \\
CompTra with BM25 & \bf 11.95 & \bf 3.64 & \bf 14.94 & \bf 4.75 & \bf 10.54 & \bf 8.89 & \bf 3.28 & \bf 15.38 & \bf 5.14 & \bf 4.30 \\
\bottomrule
\end{tabular}
}
\end{center}
\end{table*}
%
\section{Analysis}


\paragraph{Does CompTra work with weaker LLMs?} In all our experiments in Section~\ref{sec:results}, we mainly used \texttt{LLaMA 3.1 70B It}, but can CompTra work with weaker models? We compared CompTra with few-shot BM25 on FLORES~200 when both prompting approaches use the same weaker base LMs \texttt{LLaMA 3.1 8B It}, \texttt{Gemma 2 9B It} and \texttt{Command-R}. In Table~\ref{tab:scaling} (See Appendix~\ref{appendix:small} for BLEU and chrF++ scores), we observe that CompTra does work with small LMs; the average absolute performance gap across the ten FLORES languages is 1.04 MetricX with \texttt{Gemma 2 9B It} vs. 0.44 with \texttt{Gemma 2 27B It}; 1.04 with \texttt{Command-R} vs. 1.5 with \texttt{Command-R+} and 0.4 for both \texttt{LLaMAs}.
CompTra's simplicity (it does not require LMs to follow complex instructions), makes it applicable at scale.

\vspace{-0.1in}
%
\paragraph{Out-of-domain evaluation.}
In previous experiments, the selection pool shared the same domain as the evaluation set. Here, we investigate whether the gains observed disappear when it is no longer the case. To test this, we consider the setup where the evaluation set is TICO-19 test set (COVID-19; health domain) and the selection pool is FLORES~200 dev set (news domain) as opposed to the usual TICO-19 validation set (health domain). As expected and reported in Table~\ref{tab:domain}, in-domain scores are better then their out-of-domain counterparts. However, in both scenarios, applying CompTra yields gains over standalone retrieval-based few-shot MT. 
This suggests that CompTra can be successfully applied in setups where there is mismatch between the domain of the selection pool and the evaluation set.

\begin{table*}[t!]
\caption{Ablation study on the impact of the decomposition on CompTra (MetricX scores).}
\label{tab:decomposition}
\tiny
\begin{center}
\resizebox{0.9\linewidth}{!}{
\begin{tabular}{lrrrrrrrrrr}
\toprule
{} & Amharic & Burmese & Fijian & Khmer & Lao & Samoan & Sinhala & Tsonga & Turkmen & Uyghur \\
\midrule
\multicolumn{11}{l}{LLaMA 3.1 70B It} \\
\midrule
5-shot BM25    & 13.02 & 4.26 & 15.23 & 4.92 & 11.02 & 9.14 & 3.64 & 15.88 & 5.66 & 4.37 \\
CompTra        & \bf 11.95 & \bf 3.64 & \bf 14.94 & 4.75 & 10.54 & 8.89 & \bf 3.28 & \bf 15.38 & 5.14 & 4.30 \\
\hdashline
CompTra with \textbf{Words}      & 17.26 & 6.26 & 19.30 & 7.08 & 14.01 & 12.50 & 5.59 & 19.49 & 8.33 & 7.49 \\
CompTra with \textbf{Structure}  & 16.65 & 7.64 & 17.66 & 6.93 & 13.38 & 11.96 & 6.11 & 17.67 & 9.05 & 8.41 \\
CompTra with \textbf{Repeat}     & 12.96 & 4.09 & 15.25 & 4.90 & 10.87 & 9.12 & 3.53 & 15.85 & 5.59 & 4.30 \\
CompTra with \textbf{Paraphrase}    & 12.86 & 6.45 & 15.05 & \bf 4.57 & \bf 10.40 & \bf 8.40 & \bf 3.25 & 15.58 & \bf 4.64 & \bf 4.03 \\
\bottomrule
\end{tabular}
}
\end{center}
\vskip -0.15in
\end{table*}

\begin{table}[ht]
\caption{In-domain and out-of-domain evaluation (MetricX).}
\label{tab:domain}
\small
\begin{center}
\resizebox{\linewidth}{!}{
\begin{tabular}{lrrrrr}
\toprule
{} & Amharic & Khmer & Lingala & Luganda & Tamil \\
\midrule
\multicolumn{6}{c}{\textbf{\textsc{In-Domain Evaluation}}} \\
\midrule
\multicolumn{6}{l}{LLaMA 3.1 8B Instruct} \\
\midrule
5-shot BM25 & \bf 21.51 & 9.79 & 18.38 & 19.63 & \bf 3.60 \\
CompTra [with BM25] & 21.73 & \bf 9.25 & \bf 18.26 & \bf 19.24 & 3.77 \\
\midrule
\multicolumn{6}{l}{LLaMA 3.1 70B Instruct} \\
\midrule
5-shot BM25 & 12.83 & 4.54 & 14.65 & 12.81 & 2.09 \\
CompTra [with BM25] & \bf 11.97 & \bf 3.73 & \bf 14.51 & \bf 12.76 & \bf 1.52 \\
\midrule
\multicolumn{6}{c}{\textbf{\textsc{Out-of-Domain Evaluation}}} \\
\midrule
\multicolumn{6}{l}{LLaMA 3.1 8B Instruct} \\
\midrule
5-shot BM25 & \bf 23.52 & 13.42 & 20.45 & 22.38 & \bf 4.92 \\
CompTra [with BM25] & \bf 23.52 & \bf 13.10 & \bf 19.98 & \bf 21.97 & 5.26 \\
\midrule
\multicolumn{6}{l}{LLaMA 3.1 70B Instruct} \\
\midrule
5-shot BM25 & 14.96 & 5.70 & 16.86 & \bf 16.07 & 2.32 \\
CompTra [with BM25] & \bf 14.29 & \bf 5.13 & \bf 16.69 & \bf 16.03 & \bf 1.81 \\
\bottomrule
\end{tabular}
}
\end{center}
\vskip -0.2in
\end{table}

\begin{figure}[ht]
\begin{center}
    \includegraphics[width=\linewidth]{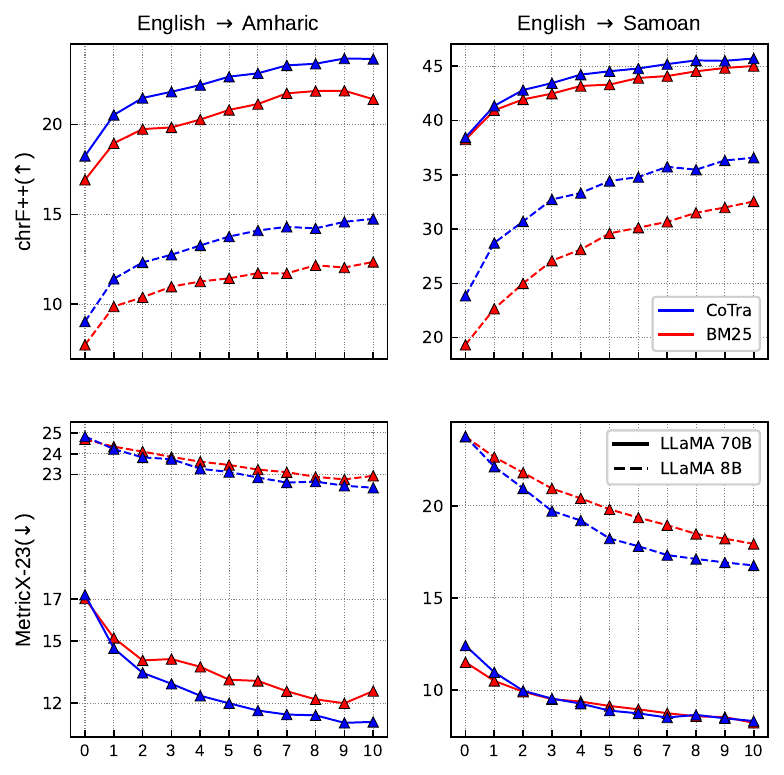}
    \vskip -0.15in
    \caption{Impact of the number of in-context examples per phrase.}
    \label{fig:scaling}
\end{center}
\end{figure}

\vspace{-0.1in}
%
\paragraph{What happens when we modify the translation step?}
In CompTra, the phrases obtained after decomposition are translated by the LLM in a few-shot manner with the help of in-context demonstrations retrieved with similarity search. In this section, we study two setups. First, we analyze the impact of the number of in-context demonstrations per phrase.
As shown in Figure~\ref{fig:scaling}, CompTra outperforms few-shot with BM25 as we vary the number of in-context demonstrations and also at scale. The performance gap is as high as 6 chrF++ and 2.5 chrF++ in Samoan and Amharic, respectively, with \texttt{LLaMA 3.1 8B It}, 1.5 MetricX in Amharic with \texttt{LLaMA 3.1 70B It}, and 1.6 MetricX in Samoan with \texttt{LLaMA 3.1 8B It}. Small values of $k$ tend to yield smaller gains and we attribute this to the fact that despite small sentences (phrases) being easier to translate for LLMs, using in-context examples helps them do it in an even better way, particularly into languages with non-Latin scripts.

Second, we replace CompTra's few-shot MT step by a direction translation obtained using a supervised translation model, here \texttt{NLLB-200-distilled-600M}~\citep{nllb2022}. The LLM therefore draws inspiration from (i.e.,~combines) the phrases' translations provided by NLLB to produce the final translation. It acts as a merger, and we call this approach \textbf{NLLB + CompTra}. We compare it to few-shot MT with BM25 and CompTra and report the results in Table~\ref{tab:nllb}.
\textbf{NLLB + CompTra} comes close or outperforms NLLB in all scenarios, proving that the translation of the phrases has a big impact on CompTra's results. \textbf{NLLB + CompTra} outperforming NLLB when the LLM few-shot performance is better than NLLB's (e.g.~Burmese, Khmer etc.) is intuitive. However, it also happens when it is not the case (Lao, Samoan) suggesting that using a strong translator to translate each subpart of a sentence and combining them with a strong ``merger'' can surpass directly using the translator on the whole sentence.

\vspace{-0.1in}
%
\paragraph{What happens when we change the retriever?}
In all our experiments, CompTra uses BM25 as the retriever. Here, we consider two different retrievers: SONAR and LCS (Longest Common Subsequence). LCS retrieval is based on the longest common subsequence between the query and the candidates after transforming them into space-separated elements. In Table~\ref{tab:retriever} we can see that BM25 is the best retriever for few-shot MT. This superiority is preserved when each retriever is used within the CompTra framework. Ultimately, BM25 is a simple, fast and performing choice.

\vspace{-0.1in}
\paragraph{What happens when we change the decomposition algorithm?}The decomposition step uses ICL and MinWikiSplit samples to break sentences into simple propositions, balancing between word-level and sentence-level for context-rich yet manageable phrases for accurate translation. We investigate four strategies: \textbf{Words}, \textbf{Repeat}, \textbf{Paraphrase} and \textbf{Structure}. \textbf{Words} decomposes a sentence into words while ignoring stop words. 
\textbf{Repeat} uses the main sentence as phrases (with $k = 4$). In the \textbf{Paraphrase} strategy, the sentence is paraphrased into at least four variations using a new \texttt{divide prompt} (See Appendix~\ref{appendix:prompts}). Finally, \textbf{Structure} divides the sentence into phrases by heuristically analyzing its dependency tree.\footnote{See appendix~\ref{appendix:structural}.} 
In Table~\ref{tab:decomposition}, we observe that \textbf{Words} and \textbf{Structure} perform the worst. A common factor in these approaches is that the phrases obtained after decomposition are not full independent sentences, making them difficult to translate. Even with sentence-translation examples in context (few-shot MT), i.e out-of-domain examples, the task remains challenging and the phrase-translation pairs obtained hurt the main MT task. \textbf{Repeat} shows marginal improvement over few-shot, indicating that phrase similarity to the main sentence matters, but repetition offers little benefit. \textbf{Paraphrase} supports this observation, outperforming \textbf{Repeat} and occasionally even slightly surpassing CompTra's native form.
It is worth noting that \textbf{Paraphrase} uses more phrases on average (4.912) than CompTra's native form (3.166) and is comparatively slower due to more tokens to generate and longer prompts.

\section{Conclusion}
We introduced a simple yet effective strategy, which we refer to as \textit{Compositional Translation} to improve the MT capabilities of LLMs. Through experiments on three MT benchmarks covering 15 different low-resource directions, we find that it outperforms the strong few-shot MT baseline with similarity search and several strong, existing strategies. It also enables smaller-scale LLMs to elicit better translation capabilities in in-domain and out-of-domain scenarios. Applying compositionality to perform MT will hopefully inspire further work on reasoning-based approaches to MT.

\section*{Acknowledgements}
This work was partly funded  by the last two authors' chairs in the PRAIRIE-PSAI institute funded by the French national agency ANR as part of the ``France 2030'' strategy under the reference ANR-23-IACL-0008. 
The authors are grateful to the OPAL infrastructure from Université Côte d'Azur for providing resources and support. This work was supported by compute credits from a Cohere For AI Research Grant, these grants are designed to support academic partners conducting research with the goal of releasing scientific artifacts and data for good projects.





\section*{Impact Statement}
This paper presents work whose goal is to advance the field of Machine Translation with Large Language Models. There are many potential societal consequences of our work, none which we feel must be specifically highlighted here.




\bibliography{bib}
\bibliographystyle{icml2025}

\newpage
\appendix
\onecolumn


\section{Additional Experiments} \label{appendix:additional_experiments}
\paragraph{MT for Nko.}
\citet{doumbouya-etal-2023-machine} extended FLORES~200 to incorporate Nko, a language spoken across multiple West African countries. We assess the ability of LLMs to use the Nko writing system, which is significantly different from the other languages we evaluate, and for which neural-based evaluation is still not the standard. Similar to FLORES~200, the selection pool contains 997 sentence pairs and the test set 1012 pairs.

Since SOTA models no longer publicly disclose the content of their training datasets, we cannot rule out the possibility that popular benchmarks might be included in these, as reported by \citet{enis2024llmnmtadvancinglowresource} with Claude 3 Opus and FLORES~200. With Nko, there is a lower risk of such a contamination, allowing for a test of CompTra in a very-low resource scenario.

We evaluate the generations with the $n$-gram matching metrics BLEU and chrF++ following~\citet{doumbouya-etal-2023-machine}. In Table~\ref{tab:nko}, we observe that CompTra works well on Nko with gains of up to 4.5 BLEU and 8 chrF++. Usually the few-shot translations contain many repeating tokens, and this issue is alleviated with the use of CompTra.

\begin{table}[ht]
\caption{Full quantitative BLEU and chrF++ results for English$\rightarrow$Nko on FLORES~200 Nko's split derived by \citet{doumbouya-etal-2023-machine}.}
\label{tab:nko}
\small
\begin{center}
\begin{tabular}{lrrrrrrrr}
\toprule
\multirow{2}{*}{Method}  & \multicolumn{2}{c}{LLaMA~3.1~70B~It} & & \multicolumn{2}{c}{Gemma~2~27B~It} & & \multicolumn{2}{c}{Command-R+}\\
\cmidrule{2-3} \cmidrule{5-6} \cmidrule{8-9}
{} & BLEU  & chrF++ &  & BLEU & chrF++ & & BLEU & chrF++ \\
\midrule
5-shot BM25 & \bf 8.80 & 19.38 &  & 10.8 & 16.49 & & 2.87 & 6.88 \\
CompTra (Ours) & 8.06 & \bf 22.16 &  & \bf 15.53 & \bf 23.04 & & \bf 8.59 & \bf 14.55 \\
\bottomrule
\end{tabular}
\end{center}
\end{table}


\paragraph{Ensembling}
CompTra outperforms few-shot MT with examples retrieved via-similarity search. In this section, we propose to do an ensembling of both approaches with the help of BLASER~2.0~QE \citep{chen-etal-2023-blaser} to account for the strengths of both approaches. Given the 2 candidate translations we choose the one with the highest quality estimation score with respect to the source sentence as the final translation. We compare this ensembling approach against each individual approach and report the results in Table~\ref{tab:blaser}.
We observe that the ensembling strategy consistently outperforms both of its individual components across all directions considered. This indicates that, while CompTra performs better than standalone few-shot MT, their outputs differ, allowing them to complement and enhance each other.

\begin{table*}[ht]
\caption{Comparison between the ensembling strategy and each of its components (MetricX scores).}
\label{tab:blaser}
\vskip 0.15in
\small
\begin{center}
\resizebox{\linewidth}{!}{
\begin{tabular}{lrrrrrrrrrr}
\toprule
{} & Amharic & Burmese & Fijian & Khmer & Lao & Samoan & Sinhala & Tsonga & Turkmen & Uyghur \\
\midrule
\multicolumn{11}{l}{LLaMA 3.1 8B Instruct} \\
\midrule
5-shot BM25 & 23.40 & 14.27 & 21.74 & 12.63 & 22.81 & 19.80 & 13.79 & 23.02 & 14.72 & 14.01 \\
CompTra (Ours) & 23.05 & 14.29 & 20.93 & 12.02 & 22.42 & 18.25 & 13.23 & 22.75 & 14.39 & 15.00 \\
Ensemble & \bf 22.65 & \bf 12.29 & \bf 20.58 & \bf 10.70 & \bf 21.72 & \bf 17.67 & \bf 11.60 & \bf 22.27 & \bf 12.41 & \bf 12.80 \\
\midrule
\multicolumn{11}{l}{LLaMA 3.1 70B Instruct} \\
\midrule
5-shot BM25 & 13.02 & 4.26 & 15.23 & 4.92 & 11.02 & 9.14 & 3.64 & 15.88 & 5.66 & 4.37 \\
CompTra (Ours) & 11.95 & 3.64 & 14.94 & 4.75 & 10.54 & 8.89 & 3.28 & 15.38 & 5.14 & 4.30 \\
Ensemble & \bf 10.93 & \bf 3.35 & \bf 14.01 & \bf 4.35 & \bf 9.62 & \bf 8.12 & \bf 2.93 & \bf 14.57 & \bf 4.60 & \bf 3.79 \\
\bottomrule
\end{tabular}
}
\end{center}
\vskip -0.1in
\end{table*}

\paragraph{Non-English-centric directions.}
We have evaluated LLMs on their ability to generate the translation of english sentences into low-resource languages. In this section, we probe them to translate from French instead. All the prompts follow the same structure as in English, with the \texttt{divide prompt} using the same sentences translated into French via Google Translate\footnote{\url{https://translate.google.com/}}. We report the results in Table~\ref{tab:french}. Translating from French is more difficult than from English as indicated by the scores. Overall, CompTra maintains its advantage over few-shot MT, though the performance gap narrows, with a few instances where few-shot MT outperforms CompTra. We attribute this to multiple factors including the quality of the \texttt{divide prompt} and the intrinsic abilities of the LLMs in French.

\begin{table*}[ht]
\caption{Full MetricX results for ten French$\rightarrow$X directions from FLORES~200.}
\label{tab:french}
\small
\begin{center}
\resizebox{\linewidth}{!}{
\begin{tabular}{lrrrrrrrrrr}
\toprule
{} & Amharic & Burmese & Fijian & Khmer & Lao & Samoan & Sinhala & Tsonga & Turkmen & Uyghur \\
\midrule
\multicolumn{11}{l}{LLaMA 3.1 8B Instruct} \\
\midrule
5-shot BM25 & 23.77 & \bf 16.22 & 22.48 & \bf 13.13 & 23.46 & 21.12 & \bf 15.58 & 23.39 & 17.12 & \bf 16.14 \\
CompTra (Ours) & \bf 23.74 & 17.41 & \bf 21.96 & 13.85 & \bf 23.10 & \bf 20.10 & 15.80 & \bf 22.97 & \bf 17.09 & 17.24 \\
\midrule
\multicolumn{11}{l}{LLaMA 3.1 70B Instruct} \\
\midrule
5-shot BM25 & 13.95 & 4.76 & 16.84 & \bf 4.96 & 11.89 & \bf 9.84 & 3.78 & \bf 17.14 & 6.15 & \bf 4.82 \\
CompTra (Ours) & \bf 13.56 & \bf 4.42 & \bf 16.68 & 5.15 & \bf 11.40 & 10.10 & \bf 3.51 & \textbf{17.09} & \bf 6.07 & 5.19 \\
\bottomrule
\end{tabular}
}
\end{center}
\vskip -0.1in
\end{table*}

\paragraph{High-resource languages as targets.} We evaluate compositional translation when translating from English to five high-resource languages: French (fra), German (deu), Spanish (spa), Portuguese (por) and Japanese (jap). As observed in Table~\ref{tab:hrl}, CompTra fails to outperform few-shot MT. We observe that zero-shot and few-shot approaches consistently perform best, typically with only a slight difference in performance between them. This partially explains the failure of CompTra, where self-generated in-context demonstrations fail to contribute meaningfully to the MT task and, in some cases, even hinder performance — similar to those retrieved via similarity search.
Additionally, we observed that smaller LMs (such as \texttt{LLaMA 3.1 8B It}) occasionally struggle to follow complex instructions included in pipelines like SBYS and TEaR, leading to poor performance. While CompTra avoids these issues due to its simplicity, it still fails to improve translation from English to other high-resource languages.

\begin{table*}[ht]
\caption{Full MetricX results for five English$\rightarrow$X high-resource directions from FLORES~200.}
\label{tab:hrl}
\small
\begin{center}
\begin{tabular}{lrrrrr}
\toprule
{} & French & German & Japanese & Portuguese & Spanish\\
\midrule
LLaMA~3.1~8B~Instruct \\
\midrule
Zero-shot & 1.49 & 1.09 & 1.39 & 1.38 & 1.41 \\
SBYS & 13.43 & 12.54 & 8.17 & 11.79 & 12.32 \\
TEaR & 8.20 & 10.65 & 12.70 & 11.21 & 9.21 \\
5-shot BM25 & \bf 1.41 & \bf 1.04 & \bf 1.30 & \bf 1.35 & \bf 1.30 \\
CompTra (Ours) & 1.65 & 1.20 & 1.75 & 1.56 & 1.52 \\
\midrule
LLaMA~3.1~70B~Instruct \\
\midrule
Zero-shot & \bf 1.03 & \bf 0.68 & 1.14 & \bf 0.99 & 1.17 \\
SBYS & 1.06 & 1.66 & 10.61 & 2.09 & 1.64 \\
TEaR & 1.19 & 0.86 & 0.97 & 1.11 & 1.17 \\
5-shot BM25 & \bf 1.02 & \bf 0.69 & \bf 0.78 & 1.04 & \bf 1.06 \\
CompTra (Ours) & 1.23 & 0.85 & 0.97 & 1.20 & 1.21 \\
\bottomrule
\end{tabular}
\end{center}
\vskip -0.1in
\end{table*}

\paragraph{Reference-free Evaluation}
While reference-based evaluation metrics are highly correlated with human judgment, they suffer from a reference bias which advantages the translation with a similar style to the reference~\citep{freitag-etal-2020-bleu}. In Table~\ref{tab:flores200} we observed that CompTra consistently outperforms 5-shot BM25 according to reference-based metrics, now we evaluate if it still holds when using the reference-free MetricX (\texttt{MetricX-23-QE-XXL}). In Table~\ref{tab:sota-qe} we observe that CompTra is still the best strategy, performing better the other across most directions. With COMETKIWI-QE (\texttt{wmt23-cometkiwi-da-xxl}; \citealp{rei2023scaling}), the conclusion is globally the same but we note a few directions where there is a  disagreement with MetricX (Samoan, Tsonga).

\begin{table*}[ht]
\caption{Full COMETKIWI-QE and MetricX-QE  results for ten English $\rightarrow$ X translation directions from FLORES~200~\citep{goyal-etal-2022-flores, nllb2022}. We compare CompTra to CoT~\citep{NEURIPS2022_8bb0d291}, MAPS~\citep{he2024exploring}, SBYS~\citep{briakou-EtAl:2024:WMT} and TEaR~\citep{feng2024tearimprovingllmbasedmachine}.}
\label{tab:sota-qe}
\small
\begin{center}
\resizebox{1.0\textwidth}{!}{
\begin{tabular}{lrrrrrrrrrrrrrr}
\toprule
\multirow{2}{*}{Methods}  & \multicolumn{2}{c}{Amharic} & & \multicolumn{2}{c}{Burmese} & & \multicolumn{2}{c}{Fijian} & & \multicolumn{2}{c}{Khmer} & & \multicolumn{2}{c}{Lao}\\
\cmidrule{2-3} \cmidrule{5-6} \cmidrule{8-9} \cmidrule{11-12} \cmidrule{14-15}
{} & {XCOMET} & {MetricX} &  & {COMET} & {MetricX} & & {COMET} & {MetricX} & & {COMET} & {MetricX} & & {XCOMET} & {MetricX}\\
\midrule
Zero-shot            & 46.66 & 11.97 &  & 77.51 & 3.08 & & 20.72 & 12.41 & & 77.45 & 3.71 & & 45.82 & 10.69 \\
Zero-shot + CoT      & 29.48 & 16.59 &  & 55.04 & 7.94 & & 17.70 & 12.18 & & 69.22 & 5.32 & & 27.24 & 16.17 \\
Zero-shot + Refine   & 48.27 & 11.68 &  & 78.56 & 2.57 & & 19.95 & 12.45 & & 78.30 & 3.44 & & 46.14 & 10.94 \\
SBYS                 & 48.43 & 10.86 &  & 78.00 & 2.91 & & \bf 21.43 & 10.74 & & 79.60 & 3.05 & & 47.10 & 9.91 \\
MAPS                 & 52.97 & 9.92  &  & 79.55 & 2.44 & & 18.26 & 13.86 & & 77.38 & 3.75 & & 49.58 & 9.78 \\
TEaR                 & 53.58 & 9.36  &  & 75.51 & 4.18 & & 20.60 & 9.77  & & 78.98 & 3.33 & & 55.66 & 7.95 \\
\hdashline
5-shot BM25          & 55.51 & 8.77  &  & 80.16 & 2.46 & & 21.19 & 8.39 & & 79.79 & 3.15 & & 58.55 & 7.07 \\
 + CoT               & 45.25 & 11.19 &  & 71.22 & 4.23 & & 19.72 & 8.57 & & 76.96 & 3.53 & & 47.00 & 9.66 \\
 + Refine            & 54.95 & 9.16  &  & \bf 80.94 & 1.98 & & 20.37 & 8.93 & & \bf 80.38 & \bf 3.03 & & 57.31 & 7.58 \\
\hdashline
CompTra (Ours)       & \bf 58.76 & \bf 7.58 &  & \bf 80.94 & \bf 1.93 & & \bf 21.38 & \bf 7.74 & & 80.13 & 3.07 & & \bf 59.64 & \bf 6.70 \\
\end{tabular}
}
\resizebox{1.0\textwidth}{!}{
\begin{tabular}{lrrrrrrrrrrrrrr}
\toprule
\multirow{2}{*}{Methods}  & \multicolumn{2}{c}{Samoan} & & \multicolumn{2}{c}{Sinhala} & & \multicolumn{2}{c}{Tsonga} & & \multicolumn{2}{c}{Turkmen} & & \multicolumn{2}{c}{Uyghur}\\
\cmidrule{2-3} \cmidrule{5-6} \cmidrule{8-9} \cmidrule{11-12} \cmidrule{14-15}
{} & {XCOMET} & {MetricX} &  & {COMET} & {MetricX} & & {COMET} & {MetricX} & & {COMET} & {MetricX} & & {XCOMET} & {MetricX}\\
\midrule
Zero-shot            & 14.68 & 7.97 &  & 75.63 & 2.72 & & 23.80 & 10.60 & & 26.49 & 5.55 & & 69.57 & 2.41 \\
Zero-shot + CoT      & 12.56 & 9.14 &  & 60.51 & 4.35 & & 21.44 & 11.83 & & 23.00 & 6.70 & & 58.29 & 4.28 \\
Zero-shot + Refine   & 14.42 & 7.95 &  & 76.12 & 2.53 & & 23.96 & 10.64 & & 27.48 & 4.71 & & 72.92 & 2.06 \\
SBYS                 & 13.64 & 7.45 &  & 77.73 & 2.19 & & 22.72 & 10.41 & & 27.15 & 4.70 & & 73.40 & 2.01 \\
MAPS                 & \bf 15.50 & 9.05 &  & 78.06 & 2.21 & & \bf 25.83 & 10.72 & & 26.24 & 6.05 & & 74.81 & 2.44 \\
TEaR                 & 15.06 & 6.69 &  & 75.15 & 3.02 & & 21.74 & 8.70  & & 28.42 & 3.64 & & 77.19 & 2.31 \\
\hdashline
5-shot BM25          & 14.60 & 6.22 &  & 78.35 & 2.33 & & 22.39 & 8.22 & & 28.47 & 3.36 & & \bf 78.92 & 1.63 \\
 + CoT               & 13.79 & 6.66 &  & 75.98 & 2.29 & & 21.62 & 8.71 & & 27.04 & 3.75 & & 72.45 & 2.28 \\
 + Refine            & 14.40 & 6.24 &  & 78.74 & 2.24 & & 22.29 & 8.21 & & \bf 28.88 & \bf 2.96 & & 79.55 & \bf 1.53 \\
\hdashline
CompTra (Ours)       & 14.93 & \bf 5.82 &  & \bf 79.39 & \bf 1.92 & & 22.60 & \bf 7.73 & & 27.81 & \bf 3.01 & & 78.16 & 1.62 \\
\bottomrule
\end{tabular}
}
\end{center}
\vskip -0.1in
\end{table*}

Moreover, we use \texttt{MetricX-23-QE-XXL} to compare how well LLMs translate the phrases compared to the main sentences. As reported in Table~\ref{tab:hypothesis}, phrases are translated more accurately, confirming CompTra's core hypothesis. Heuristically, we observed that a larger quality gap between phrase translations and main sentence translations correlates with better CompTra performance. However, with the \textbf{Paraphrase} strategy, we observe that \texttt{LLaMA 3.1 70B It} better translate a self-generated paraphrase a sentence than the sentence itself but not as well as the short phrases obtained with the native \texttt{divide prompt}. Indeed, how good phrases are translated matters but the similarity the semantic similarity between sentence and the phrases seems to be more important for the success of CompTra.

\begin{table*}[ht]
\caption{Full MetricX-QE results for ten English$\rightarrow$ X directions. We compare how accurately phrases are translated compared to main sentences.}
\label{tab:hypothesis}
\small
\begin{center}
\resizebox{\linewidth}{!}{
\begin{tabular}{lrrrrrrrrrr}
\toprule
{} & Amharic & Burmese & Fijian & Khmer & Lao & Samoan & Sinhala & Tsonga & Turkmen & Uyghur \\
\midrule
\multicolumn{11}{l}{LLaMA 3.1 70B It} \\
\midrule
5-shot BM25 & 8.77 & 2.46 & 8.39 & 3.15 & 7.07 & 6.22 & 2.33 & 8.22 & 3.36 & 1.63 \\
\hdashline
Phrases & 3.89 & 0.88 & 5.47 & 1.47 & 3.34 & 2.95 & 0.88 & 4.97 & 2.02 & 1.08 \\
CompTra & 7.58 & 1.93 & 7.74 & 3.07 & 6.70 & 5.82 & 1.92 & 7.73 & 3.01 & 1.62 \\
\hdashline
\textbf{Paraphrase}'s phrases & 8.45 & 2.15 & 8.22 & 2.66 & 6.34 & 5.82 & 1.83 & 8.20 & 2.95 & 1.41 \\
CompTra with \textbf{Paraphrase} & 8.41 & 4.94 & 7.72 & 2.92 & 6.56 & 5.46 & 1.91 & 7.63 & 2.52 & 1.34 \\
\midrule
\multicolumn{11}{l}{Gemma 2 27B It} \\
\midrule
Phrases    & 4.65 & 1.99 & 7.56 & 2.11 & 3.04 & 5.48 & 1.84 & 5.45 & 1.95 & 2.18 \\
CompTra & 8.19 & 3.86 & 10.63 & 4.23 & 5.80 & 9.88 & 3.68 & 8.04 & 2.67 & 3.32 \\
\midrule
\multicolumn{11}{l}{Command-R+} \\
\midrule
Phrases    & 14.09 & 4.79 & 8.09 & 8.38 & 7.15 & 8.87 & 2.53 & 10.13 & 1.77 & 1.79 \\
CompTra  & 19.88 & 8.39 & 10.74 & 12.28 & 12.42 & 13.85 & 4.81 & 15.14 & 2.45 & 2.91 \\
\bottomrule
\end{tabular}
}
\end{center}
\vskip -0.1in
\end{table*}
\vspace{-0.1in}

\section{Additional Results}\label{appendix:additional_results}

\subsection{BLEU and chrF++ results on FLORES~200} \label{appendix:bleu-chrf-flores}

As mentioned previously we additionally present results with BLEU and chrF++ scores of CompTra against baselines in Table~\ref{tab:flores200-bleu-chrf}) for transparency reasons. The results show the same pattern as the XCOMET and MetricX results shown in the main part of the paper. CompTra outperforms few-shot with SONAR and BM25 in all scenarios. When it comes to LRLs, few-shot MT with example selection via similarity search should be the standard as it always outperforms zero-shot MT and has been proven to perform better than random selection~\citep{zebaze2024incontextexampleselectionsimilarity}.

\begin{table*}[ht]
\caption{Full BLEU and chrF++ results for ten English$\xrightarrow{}$X directions from FLORES~200~\citep{goyal-etal-2022-flores, nllb2022}.}
\label{tab:flores200-bleu-chrf}
\vskip 0.15in
\small
\begin{center}
\resizebox{1.0\textwidth}{!}{
\begin{tabular}{lrrrrrrrrrrrrrr}
\toprule
\multirow{2}{*}{Methods}  & \multicolumn{2}{c}{Amharic} & & \multicolumn{2}{c}{Burmese} & & \multicolumn{2}{c}{Fijian} & & \multicolumn{2}{c}{Khmer} & & \multicolumn{2}{c}{Lao}\\
\cmidrule{2-3} \cmidrule{5-6} \cmidrule{8-9} \cmidrule{11-12} \cmidrule{14-15}
{} & {BLEU} & {chrF++} &  & {BLEU} & {chrF++} & & {BLEU} & {chrF++} & & {BLEU} & {chrF++} & & {BLEU} & {chrF++}\\
\midrule
LLaMA~3~70B~It \\
\midrule
Zero-shot     & 8.81 & 16.95 &  & 17.41  & 34.78 & & 7.70 & 28.05 & & 18.46 & 30.81 & & 8.30 & 24.52 \\
5-shot SONAR  & 11.24 & 20.25 &  & 19.05 & 36.22 & & 11.85 & 34.98 & & 20.36 & 33.43 & & 14.09 & 31.96 \\
5-shot BM25   & 11.86 & 20.92 &  & 19.49 & 36.80 & & 12.19 & 35.34 & & 20.14 & 37.53 & & 15.13 & 32.70 \\
CompTra (Ours)  & \bf 12.64 & \bf 22.67 &  & \bf 19.84 & \bf 38.03 & & \bf 12.94 & \bf 38.02 & & \bf 20.60 & \bf 33.06 & & \bf 16.37 & \bf 33.14 \\
\midrule
Gemma~2~27B~It \\
\midrule
Zero-shot     & 7.85 & 17.24 &  & 10.70 & 28.50 & & 7.39 & 28.57 & & 11.30 & 25.13 & & 8.91 & 25.05 \\
5-shot SONAR  & 10.19 & 19.86 &  & 13.44 & 31.72 & & 9.97 & 31.74 & & 14.36 & 29.16 & & 15.14 & 33.75 \\
5-shot BM25   & 10.52 & 20.24 &  & 13.98 & 32.08 & & 9.93 & 31.88 & & 14.57 & 28.86 & & 15.58 & 33.73 \\
CompTra (Ours) & \bf 11.67 & \bf 21.46 &  & \bf 15.11 & \bf 32.99 & & \bf 11.63 & \bf 35.72 & & \bf 15.61 & \bf 29.52 & & \bf 17.20 & \bf 34.62\\
\midrule
Command-R+ \\
\midrule
Zero-shot    & 2.59 & 8.21  &  & 5.32 & 21.60 & & 4.36 & 22.45 & & 5.96 & 19.47 & & 4.15 & 21.04 \\
5-shot SONAR & 4.40 & 10.69 &  & 9.40 & 27.18 & & 8.38 & 29.74 & & 8.32 & 22.18 & & 8.72 & 27.21 \\
5-shot BM25  & 4.65 & 11.17 &  & 9.70 & 27.63 & & 8.94 & 30.30 & & 8.54 & 21.77 & & \bf 9.30 & 27.54 \\
CompTra (Ours) & \bf 6.66 & \bf 14.56 &  & \bf 12.30 & \bf 30.73 & & \bf 11.20 & \bf 36.36 & & \bf 10.99 & \bf 24.97 & & 9.10 & \bf 28.72\\
\end{tabular}
}
\resizebox{1.0\textwidth}{!}{
\begin{tabular}{lrrrrrrrrrrrrrr}
\toprule
\multirow{2}{*}{Methods}  & \multicolumn{2}{c}{Samoan} & & \multicolumn{2}{c}{Sinhala} & & \multicolumn{2}{c}{Tsonga} & & \multicolumn{2}{c}{Turkmen} & & \multicolumn{2}{c}{Uyghur}\\
\cmidrule{2-3} \cmidrule{5-6} \cmidrule{8-9} \cmidrule{11-12} \cmidrule{14-15}
{} & {BLEU} & {chrF++} &  & {BLEU} & {chrF++} & & {BLEU} & {chrF++} & & {BLEU} & {chrF++} & & {BLEU} & {chrF++}\\
\midrule
LLaMA~3~70B~It \\
\midrule
Zero-shot      & 16.04 & 38.25 &  & 23.71 & 36.20 & & 6.34 & 25.34 & & 13.40 & 32.69 & & 11.51 & 25.04 \\
5-shot SONAR   & 20.93 & 42.90 &  & 25.34 & 38.03 & & 10.28 & 31.96 & & 17.99 & 38.01 & & 19.41 & 37.05 \\
5-shot BM25    & 21.47 & 43.29 &  & 25.40 & 38.02 & & 11.13 & 33.11 & & 18.56 & 38.66 & & 20.01 & 37.53 \\
CompTra (Ours) & \bf 21.59 & \bf 44.53 &  & \bf 26.20 & \bf 39.67 & & \bf 11.55 & \bf 35.01 & & \bf 19.69 & \bf 40.64 & & \bf 21.11 & \bf 39.00 \\
\midrule
Gemma~2~27B~It \\
\midrule
Zero-shot & 10.63 & 32.83 &  & 14.75 & 27.41 & & 6.77 & 27.25 & & 10.17 & 30.21 & & 6.27 & 22.62 \\
5-shot SONAR & 13.87 & 36.35 &  & 17.96 & 30.61 & & 10.22 & 32.34 & & 13.85 & 35.03 & & 11.07 & 28.08 \\
5-shot BM25 & 14.25 & 36.77 &  & 18.57 & 31.01 & & 10.90 & 33.13 & & 14.85 & 35.60 & & 11.90 & 28.55 \\
CompTra (Ours) & \bf 15.34 & \bf 38.70 &  & \bf 20.24 & \bf 32.67 & & \bf 11.65 & \bf 35.07 & & \bf 16.11 & \bf 36.78 & & \bf 13.32 & \bf 30.47\\
\midrule
Command-R+ \\
\midrule
Zero-shot    & 5.18  & 19.90 &  & 12.98 & 26.80 & & 3.19 & 14.19 & & 13.94 & 34.17 & & 7.48  & 23.10 \\
5-shot SONAR & 10.96 & 28.24 &  & 16.56 & 30.31 & & 5.73 & 21.61 & & 18.64 & 39.08 & & 13.26 & 29.93 \\
5-shot BM25  & 12.19 & 29.66 &  & 17.45 & 31.02 & & 6.24 & 22.68 & & 19.69 & 39.85 & & 14.74 & 31.57 \\
CompTra (Ours) & \bf 14.67 & \bf 36.37 &  & \bf 19.64 & \bf 33.71 & & \bf 7.84 & \bf 27.80 & & \bf 20.72 & \bf 41.09 & & \bf 16.59 & \bf 33.77 \\
\bottomrule
\end{tabular}
}
\end{center}
\vskip -0.1in
\end{table*}

Moreover, as reported in Table~\ref{tab:sota-bleu-chrf}, CompTra has the interesting property of improving both neural-based and string-matching metrics as opposed to existing strategies. The superiority of CompTra is observed across four distinct metrics (XCOMET, MetricX, BLEU and chrF++), each with unique properties, highlighting its robustness.

\begin{table*}[ht]
\caption{Full BLEU and chrF++ results for ten English$\rightarrow$X directions from FLORES~200~\citep{goyal-etal-2022-flores, nllb2022}. We compare CompTra to CoT~\citep{NEURIPS2022_8bb0d291}, MAPS~\citep{he2024exploring}, SBYS~\citep{briakou-EtAl:2024:WMT} and TEaR~\citep{feng2024tearimprovingllmbasedmachine}.}
\label{tab:sota-bleu-chrf}
\small
\begin{center}
\resizebox{1.0\textwidth}{!}{
\begin{tabular}{lrrrrrrrrrrrrrr}
\toprule
\multirow{2}{*}{Methods}  & \multicolumn{2}{c}{Amharic} & & \multicolumn{2}{c}{Burmese} & & \multicolumn{2}{c}{Fijian} & & \multicolumn{2}{c}{Khmer} & & \multicolumn{2}{c}{Lao}\\
\cmidrule{2-3} \cmidrule{5-6} \cmidrule{8-9} \cmidrule{11-12} \cmidrule{14-15}
{} & {BLEU} & {chrF++} &  & {BLEU} & {chrF++} & & {BLEU} & {chrF++} & & {BLEU} & {chrF++} & & {BLEU} & {chrF++}\\
\midrule
Zero-shot            & 8.81 & 16.95 &  & 17.41  & 34.78 & & 7.70 & 28.05 & & 18.46 & 30.81 & & 8.30 & 24.52 \\
Zero-shot + CoT      & 6.64 & 13.60 &  & 11.42 & 29.00 & & 6.48 & 26.86 & & 15.13 & 27.98 & & 3.61 & 16.87 \\
Zero-shot + Refine   & 8.53 & 16.65 &  & 16.91 & 34.02 & & 7.70 & 28.41 & & 17.79 & 29.94 & & 7.60 & 22.97 \\
SBYS                 & 8.76 & 17.44 &  & 16.43 & 34.12 & & 7.92 & 30.04 & & 17.62 & 29.95 & & 8.18 & 25.31 \\
MAPS                 & 9.47 & 17.95 &  & 17.42 & 34.57 & & 6.25 & 23.41 & & 18.48 & 31.66 & & 9.21 & 25.37 \\
TEaR                 & 11.19 & 20.04 &  & 17.93 & 34.41 & & 11.12 & 34.15 & & 19.56 & 32.30 & & 13.62 & 30.84 \\
\hdashline
5-shot BM25          & 11.86 & 20.92 &  & 19.49 & 36.80 & & 12.19 & 35.34 & & 20.14 & \bf 33.15 & & 15.13 & 32.70 \\
 + CoT               & 10.56 & 19.13 &  & 17.06 & 34.26 & & 10.67 & 34.47 & & 18.47 & 31.29 & & 11.29 & 27.81 \\
 + Refine            & 11.05 & 19.82 &  & 18.66 & 35.85 & & 11.51 & 34.76 & & 18.80 & 31.10 & & 13.36 & 29.65 \\
\hdashline
CompTra (Ours)         & \bf 12.64 & \bf  22.67 &  & \bf 19.84 & \bf 38.03 & & \bf 12.94 & \bf 38.02 & & \bf 20.60 & 33.06 & & \bf 16.37 & \bf 33.14 \\
\end{tabular}
}
\resizebox{1.0\textwidth}{!}{
\begin{tabular}{lrrrrrrrrrrrrrr}
\toprule
\multirow{2}{*}{Methods}  & \multicolumn{2}{c}{Samoan} & & \multicolumn{2}{c}{Sinhala} & & \multicolumn{2}{c}{Tsonga} & & \multicolumn{2}{c}{Turkmen} & & \multicolumn{2}{c}{Uyghur}\\
\cmidrule{2-3} \cmidrule{5-6} \cmidrule{8-9} \cmidrule{11-12} \cmidrule{14-15}
{} & {BLEU} & {chrF++} &  & {BLEU} & {chrF++} & & {BLEU} & {chrF++} & & {BLEU} & {chrF++} & & {BLEU} & {chrF++}\\
\midrule
Zero-shot            & 16.04 & 38.25 &  & 23.71 & 36.20 & & 6.34 & 25.34 & & 13.40 & 32.69 & & 11.51 & 25.04 \\
Zero-shot + CoT      & 13.66 & 35.55 &  & 19.53 & 32.36 & & 5.26 & 24.05 & & 11.68 & 31.12 & & 12.14 & 27.53 \\
Zero-shot + Refine   & 15.82 & 37.98 &  & 22.79 & 35.19 & & 6.20 & 25.26 & & 13.64 & 33.14 & & 11.86 & 25.48 \\
SBYS                 & 14.85 & 37.54 &  & 22.91 & 36.00 & & 6.14 & 26.16 & & 13.18 & 33.02 & & 14.80 & 31.76 \\
MAPS                 & 15.02 & 36.20 &  & 23.16 & 35.45 & & 5.89 & 23.37 & & 11.75 & 29.47 & & 14.28 & 29.41 \\
TEaR                 & 19.71 & 42.02 &  & 24.67 & 36.76 & & 9.88 & 31.74 & & 16.53 & 36.97 & & 18.60 & 36.18 \\
\hdashline
5-shot BM25          & 21.47 & 43.29 &  & 25.40 & 38.02 & & 11.13 & 33.11 & & 18.56 & 38.66 & & 20.01 & 37.53 \\
 + CoT               & 17.72 & 40.06 &  & 24.68 & 37.37 & & 9.51 & 31.50 & & 17.22 & 37.30 & & 19.38 & 36.25 \\
 + Refine            & 19.03 & 41.45 &  & 24.19 & 36.44 & & 10.24 & 32.18 & & 17.57 & 37.51 & & 19.08 & 36.17 \\
\hdashline
CompTra (Ours)       & \bf 21.59 & \bf 44.53 &  & \bf 26.20 & \bf 39.67 & & \bf 11.55 & \bf 35.01 & & \bf 19.69 & \bf 40.64 & & \bf 21.11 & \bf 39.00\\
\bottomrule
\end{tabular}
}
\end{center}
\vskip -0.1in
\end{table*}

\subsection{BLEU and chrF++ results on NTREX~128} \label{appendix:bleu-chrf-ntrex}

We present results with BLEU and chrF++ scores of CompTra against baselines on NTREX 128 in Table~\ref{tab:ntrex-bleu-chrf} and TICO-19 in Table~\ref{tab:tico-bleu-chrf}. The results are the same as in Table~\ref{tab:ntrex} and Table~\ref{tab:tico} when CompTra outperforms few-shot MT with BM25 and SONAR.

\begin{table*}[ht]
\caption{Full BLEU and chrF++ results for ten English$\xrightarrow{}$X directions from NTREX~128~\citep{federmann-etal-2022-ntrex, barrault-etal-2019-findings}.}
\label{tab:ntrex-bleu-chrf}
\small
\begin{center}
\resizebox{1.0\textwidth}{!}{
\begin{tabular}{lrrrrrrrrrrrrrr}
\toprule
\multirow{2}{*}{Methods}  & \multicolumn{2}{c}{Amharic} & & \multicolumn{2}{c}{Fijian} & & \multicolumn{2}{c}{Shona} & & \multicolumn{2}{c}{Somali} & & \multicolumn{2}{c}{Tswana}\\
\cmidrule{2-3} \cmidrule{5-6} \cmidrule{8-9} \cmidrule{11-12} \cmidrule{14-15}
{} & {BLEU} & {chrF++} &  & {BLEU} & {chrF++} & & {BLEU} & {chrF++} & & {BLEU} & {chrF++} & & {BLEU} & {chrF++}\\
\midrule
LLaMA~3.1~70B~It \\
\midrule
5-shot BM25 & 8.11 & 15.28 &  & 12.42 & 35.59 & & 11.60 & 31.67 & & 12.76 & 37.12 & & 18.67 & 39.87 \\
CompTra (Ours) & \bf 9.13 & \bf 16.71 &  & \bf 13.42 & \bf 38.30 & & \bf 11.87 & \bf 33.92 & & \bf 13.21 & \bf 38.40 & & \bf 20.07 & \bf 41.61\\
\midrule
Gemma~2~27B~It \\
\midrule
5-shot BM25 & 6.99 & 15.16 &  & 10.25 & 32.34 & & 12.78 & 35.33 & & 12.65 & 37.11 & & 18.66 & 40.83 \\
CompTra (Ours) & \bf 8.33 & \bf 16.55 &  & \bf 12.48 & \bf 36.53 & & \bf 13.55 & \bf 36.46 & & \bf 13.27 & \bf 37.69 & & \bf 19.84 & \bf 42.19 \\
\midrule
Command-R+ \\
\midrule
5-shot BM25 & 3.09 & 8.82 &  & 9.46 & 30.37 & & 7.50 & 25.08 & & 8.45 & 28.14 & & 11.77 & 29.57 \\
CompTra (Ours) & \bf 4.61 & \bf 11.38 &  & \bf 11.92 & \bf 36.21 & & \bf 9.63 & \bf 29.72 & & \bf 9.93 & \bf 32.09 & & \bf 14.98 & \bf 35.82\\
\bottomrule
\end{tabular}
}
\end{center}
\end{table*}

\begin{table*}[ht]
\caption{Full BLEU and chrF++ results for 5 English$\rightarrow$X directions from TICO-19 \citep{anastasopoulos-etal-2020-tico}.}
\small
\label{tab:tico-bleu-chrf}
\begin{center}
\resizebox{1.0\textwidth}{!}{
\begin{tabular}{lrrrrrrrrrrrrrr}
\toprule
\multirow{2}{*}{Methods}  & \multicolumn{2}{c}{Amharic} & & \multicolumn{2}{c}{Khmer} & & \multicolumn{2}{c}{Lingala} & & \multicolumn{2}{c}{Luganda} & & \multicolumn{2}{c}{Tamil}\\
\cmidrule{2-3} \cmidrule{5-6} \cmidrule{8-9} \cmidrule{11-12} \cmidrule{14-15}
{} & {BLEU} & {chrF++} &  & {BLEU} & {chrF++} & & {BLEU} & {chrF++} & & {BLEU} & {chrF++} & & {BLEU} & {chrF++}\\
\midrule
LLaMA~3.1~70B~It \\
\midrule
5-shot BM25 & 11.90 & 20.90 &  & 32.94 & 44.08 & & 14.47 & 35.63 & & 15.83 & 36.00 & & 32.05 & 50.43 \\
CompTra (Ours) & \bf 13.44 & \bf 23.08 &  & \bf 34.71 & \bf 45.60 & & \bf 15.17 & \bf 39.21 & & \bf 16.21 & \bf 37.86 & & \bf 33.60 & \bf 52.02 \\
\midrule
Gemma~2~27B~It \\
\midrule
5-shot BM25 & 10.95 & 21.25 &  & 26.56 & 39.96 & & 15.27 & 37.51 & & 14.00 & 33.77 & & 27.56 & 47.53 \\
CompTra (Ours) & \bf 12.50 & \bf 22.81 &  & \bf 28.67 & \bf 41.40 & & \bf 16.11 & \bf 39.91 & & \bf 15.77 & \bf 36.52 & & \bf 28.62 & \bf 48.21 \\
\midrule
Command-R+ \\
\midrule
5-shot BM25 & 5.90 & 12.98 &  & 18.34 & 31.18 & & 11.64 & 31.00 & & 8.54 & 24.56 & & 28.77 & 47.62 \\
CompTra (Ours) & \bf 8.25 & \bf 16.74 &  & \bf 20.76 & \bf 34.11 & & \bf 14.19 & \bf 36.86 & & \bf 11.18 & \bf 30.05 & & \bf 29.63 & \bf 48.69\\
\bottomrule
\end{tabular}
}
\end{center}
\vskip -0.1in
\end{table*}

\subsection{BLEU and chrF++ results on FLORES 200 with small LMs} \label{appendix:small}

We reported that CompTra works very well with smaller LMs in Table~\ref{tab:scaling} by reporting the MetricX scores. In Table~\ref{tab:scaling-bleu-chrf}, we show that the performance gains provided by CompTra are also observable in terms of BLEU and chrF++. Moreover, we compare SBYS, TEaR, MAPS and 5-shot BM25 + Self-refine to CompTra using \texttt{LLaMA 3.1 8B It}, \texttt{Gemma 2 9B It} and \texttt{Command-R} and report the results in Table~\ref{tab:sota-small}. CompTra ends up being the best approach at this scale too. The models sometime struggle to directly refine their answers, leading the performances of 5-shot BM25 + Self-refine to be worse or equal to 5-shot BM25. SBYS does not work well with \texttt{LLaMA 3.1 8B It} but give good results with \texttt{Gemma 2 9B It} (Similar to the strong results they achieved with Gemini; \citealp{geminiteam2024gemini15unlockingmultimodal}), outperforming CompTra in some scenarios. With \texttt{Command-R}, SBYS does not fail as it does with \texttt{LLaMA 3.1 8B It} but it remains worse than CompTra in most scenarios.

\begin{table*}[ht]
\caption{Full quantitative BLEU and chrF++ for ten English$\xrightarrow{}$X directions from FLORES~200~\citep{goyal-etal-2022-flores, nllb2022} with small LMs.}
\label{tab:scaling-bleu-chrf}
\vskip 0.15in
\small
\begin{center}
\resizebox{1.0\textwidth}{!}{
\begin{tabular}{lrrrrrrrrrrrrrr}
\toprule
\multirow{2}{*}{Methods}  & \multicolumn{2}{c}{Amharic} & & \multicolumn{2}{c}{Burmese} & & \multicolumn{2}{c}{Fijian} & & \multicolumn{2}{c}{Khmer} & & \multicolumn{2}{c}{Lao}\\
\cmidrule{2-3} \cmidrule{5-6} \cmidrule{8-9} \cmidrule{11-12} \cmidrule{14-15}
{} & {BLEU} & {chrF++} &  & {BLEU} & {chrF++} & & {BLEU} & {chrF++} & & {BLEU} & {chrF++} & & {BLEU} & {chrF++}\\
\midrule
LLaMA~3.1~8B~It \\
\midrule
5-shot BM25   & 4.27 & 11.37 &  & 10.09 & 28.09 & & 7.37 & 25.92 & & 10.44 & 25.17 & & 4.17 & 18.84 \\
CompTra (Ours)  & \bf 5.76 & \bf 13.73 &  & \bf 11.62 & \bf 29.88 & & \bf 8.60 & \bf 31.32 & & \bf 11.68 & \bf 26.30 & & \bf 3.53 & \bf 19.74 \\
\midrule
Gemma~2~9B~It \\
\midrule
5-shot BM25   & 8.34 & 17.66 &  & 9.69 & 27.91 & & 8.70 & 29.26 & & 10.66 & 25.57 & & 10.77 & 28.79 \\
CompTra (Ours)  & \bf 9.86 & \bf 19.66 &  & \bf 11.35 & \bf 30.05 & & \bf 10.61 & \bf 34.13 & & \bf 12.09 & \bf 26.95 & & \bf 13.64 & \bf 32.07 \\
\midrule
Command-R \\
\midrule
5-shot BM25   & 2.66 & 8.85 &  & 5.69 & 22.48 & & 7.83 & 29.25 & & 5.31 & 19.21 & & \bf 4.51 & 20.65 \\
CompTra (Ours)  & \bf 4.13 & \bf 12.02 &  & \bf 8.50 & \bf 27.09 & & \bf 8.24 & \bf 34.38 & & \bf 7.06 & \bf 20.90 & & 4.47 & \bf 22.48\\
\end{tabular}
}
\resizebox{1.0\textwidth}{!}{
\begin{tabular}{lrrrrrrrrrrrrrr}
\toprule
\multirow{2}{*}{Methods}  & \multicolumn{2}{c}{Samoan} & & \multicolumn{2}{c}{Sinhala} & & \multicolumn{2}{c}{Tsonga} & & \multicolumn{2}{c}{Turkmen} & & \multicolumn{2}{c}{Uyghur}\\
\cmidrule{2-3} \cmidrule{5-6} \cmidrule{8-9} \cmidrule{11-12} \cmidrule{14-15}
{} & {BLEU} & {chrF++} &  & {BLEU} & {chrF++} & & {BLEU} & {chrF++} & & {BLEU} & {chrF++} & & {BLEU} & {chrF++}\\
\midrule
LLaMA~3.1~8B~It \\
\midrule
5-shot BM25   & 12.01 & 29.53 &  & 12.69 & 24.67 & & 5.53 & 21.07 & & 8.93 & 26.22 & & 10.65 & 27.20 \\
CompTra (Ours)  & \bf 13.95 & \bf 34.46 &  & \bf 14.84 & \bf 27.71 & & \bf 7.43 & \bf 27.53 & & \bf 9.80 & \bf 29.38 & & \bf 11.90 & \bf 29.15\\
\midrule
Gemma~2~9B~It \\
\midrule
5-shot BM25   & 14.98 & 34.59 &  & 16.04 & 29.99 & & 6.86 & 23.70 & & 9.92 & 29.33 & & 5.83 & 18.25 \\
CompTra (Ours)  & \bf 16.77 & \bf 38.42 &  & \bf 18.09 & \bf 31.77 & & \bf 8.71 & \bf 29.54 & & \bf 11.24 & \bf 31.56 & & \bf 9.14 & \bf 24.32\\
\midrule
Command-R \\
\midrule
5-shot BM25   & 8.88 & 25.48 &  & 10.21 & 23.44 & & 5.92 & 22.42 & & 13.62 & 33.88 & & 9.04 & 24.30 \\
CompTra (Ours)  & \bf 11.84 & \bf 33.18 &  & \bf 13.36 & \bf 27.81 & & \bf 6.60 & \bf 27.68 & & \bf 14.83 & \bf 35.60 & & \bf 10.67 & \bf 28.50 \\
\bottomrule
\end{tabular}
}
\end{center}
\vskip -0.1in
\end{table*}

\begin{table*}[ht]
\caption{Full MetricX results for ten English $\rightarrow$ X translation directions from FLORES~200 with small LMs. We compare CompTra to Self-refine~\citep{chen2024iterativetranslationrefinementlarge}, MAPS~\citep{he2024exploring}, SBYS~\citep{briakou-EtAl:2024:WMT} and TEaR~\citep{feng2024tearimprovingllmbasedmachine}.}
\label{tab:sota-small}
\small
\begin{center}
\resizebox{\linewidth}{!}{
\begin{tabular}{lrrrrrrrrrr}
\toprule
{} & Amharic & Burmese & Fijian & Khmer & Lao & Samoan & Sinhala & Tsonga & Turkmen & Uyghur \\
\midrule
\multicolumn{11}{l}{LLaMA 3.1 8B Instruct}  \\
\midrule
MAPS        & 24.38 & 15.49 & 23.28 & 13.50 & 24.53 & 22.96 & 13.39 & 23.94 & 12.54 & 16.45 \\
SBYS        & 24.76 & 22.14 & 24.66 & 16.55 & 24.92 & 24.51 & 23.25 & 24.67 & 22.10 & 22.36 \\
TEaR        & 24.05 & 16.98 & 22.71 & 14.05 & 23.51 & 20.95 & 16.48 & 23.64 & 17.98 & 17.46 \\
5-shot BM25 & 23.40 & 14.27 & 21.74 & 12.63 & 22.81 & 19.80 & 13.79 & 23.02 & 14.72 & \bf 14.01 \\
+ Refine    & 23.54 & \bf 14.23 & 22.34 & 14.00 & 23.76 & 20.77 & 14.16 & 23.20 & \bf 14.18 & 14.46 \\
CompTra     & \bf 23.06 & 14.29 & \bf 20.93 & \bf 12.02 & \bf 22.41 & \bf 18.25 & \bf 13.23 & \bf 22.75 & 14.39 & 15.00 \\
\midrule
\multicolumn{11}{l}{Gemma 2 9B It} \\
\midrule
MAPS        & 15.42 & 14.35 & 21.86 & 12.40 & 14.28 & 20.57 & 8.98 & 22.14 & 5.66 & 22.38 \\
SBYS        & \bf 15.04 & 15.30 & \bf 18.81 & 12.18 & 13.97 & \bf 15.92 & 11.01 & \bf 18.34 & \bf 5.01 & 20.06 \\
TEaR        & 15.75 & 13.21 & 20.54 & 11.30 & 14.52 & 17.14 & \bf 8.70 & 21.60 & 8.55 & 20.96 \\
5-shot BM25 & 15.99 & 13.05 & 20.66 & 11.92 & 15.21 & 17.61 & 9.13 & 20.99 & 8.36 & 21.07 \\
+ Refine    & 15.64 & 13.30 & 20.93 & 11.73 & 15.04 & 17.71 & \bf 8.73 & 21.56 & 5.19 & 20.89 \\
CompTra     & 15.66 & \bf 12.31 & 19.63 & \bf 11.23 & \bf 13.67 & \bf 15.93 & 8.82 & 19.82 & 7.69 & \bf 19.19 \\
\midrule
\multicolumn{11}{l}{Command R} \\
\midrule
MAPS & 24.87 & 22.28 & 23.85 & 23.09 & 23.79 & 23.58 & 15.15 & 24.36 & 9.61 & 16.95 \\
SBYS & \bf 23.57 & 23.09 & 22.98 & 23.67 & 23.68 & 22.68 & 16.78 & 22.26 & 7.57 & 20.43 \\
TEaR & 24.57 & 21.77 & 21.39 & 21.73 & 22.87 & 21.65 & 15.94 & 22.70 & 7.38 & 16.53 \\
5-shot BM25 & 24.38 & 20.94 & 21.24 & 21.64 & 22.68 & 21.67 & 15.50 & 22.46 & 7.01 & 16.36 \\
+ Refine & 24.57 & 21.35 & 22.30 & 21.78 & 23.08 & 22.78 & 15.21 & 23.46 & 6.60 & 15.99 \\
CompTra & 24.39 & \bf 19.33 & \bf 20.59 & \bf 20.48 & \bf 21.88 & \bf 20.82 & \bf 12.91 & \bf22.16 & \bf 5.95 & \bf 14.99 \\
\bottomrule
\end{tabular}
}
\end{center}
\vskip -0.1in
\end{table*}


\clearpage
\section{Implementation details}
\subsection{General remarks} \label{appendix:general_remarks}
When translating into LRLs, particularly languages with non-Latin scripts, it is important to generate the right amount of tokens. Current tokenizers tend to require more tokens for non-Latin scripts, thus translating a 100-token English sentence in French can use half as much tokens as doing so in Amharic. This is the reason why we set \verb|max_new_tokens| to 500. However, it comes with the risk of overgeneration~\citep{bawden-yvon-2023-investigating, zebaze2024incontextexampleselectionsimilarity}. It occurs when translating into LRLs with base models but also with intruction fine-tuned/chat models and usually take the form of repeating $n$-grams at the end of the generations. This is done by space-separating the generation, identifying a bigram which occurs more than eight times and drop the rest of the sentence after its first occurrence. For the statistical significant comparison between each pair strategies, we follow \citep{koehn-2004-statistical} and use paired bootstrap resampling with 300 samples of 500 sentences and a $p$-value threshold of 0.05.

In this paper, a phrase is to be understood as contiguous subpart of a sentence, in the context of phrase-based MT and not in the linguistic sense of a phrase or constituent.

\subsection{Models, Datasets and Tools}

In Table~\ref{tab:url}, we list the links to the relevant resources used for experiments.
\begin{table*}[!ht]
    \centering\small
    \caption{Links to datasets, benchmarks and models.}
    \resizebox{\linewidth}{!}{
    \begin{tabular}{l l}
    \toprule
    \multicolumn{2}{c}{\textit{Datasets}} \\
    \midrule
    FLORES~200 & \url{https://huggingface.co/datasets/facebook/flores} \\
    Machine Translation for Nko & \url{https://github.com/common-parallel-corpora/common-parallel-corpora} \\
    NTREX & \url{https://github.com/MicrosoftTranslator/NTREX/tree/main} \\
    NTREX HF & \url{hhttps://huggingface.co/datasets/mteb/NTREX} \\
    TICO-19 & \url{https://huggingface.co/datasets/gmnlp/tico19}\\
    \midrule
    \multicolumn{2}{c}{\textit{Models evaluated}} \\
    \midrule
    Command-R & \texttt{command-r-08-2024} \\
    Command-R+ & \texttt{command-r-plus-08-2024} \\
    Gemma~2~2B~It & \url{https://huggingface.co/google/gemma-2-2b-it} \\
    Gemma~2~9B~It & \url{https://huggingface.co/google/gemma-2-9b-it} \\
    Gemma~2~27B~It & \url{https://huggingface.co/google/gemma-2-27b-it} \\
    LLaMA~3.1~8B~It & \url{https://huggingface.co/meta-llama/Meta-Llama-3.1-8B-Instruct} \\
    LLaMA~3.1~70B~It & \url{https://huggingface.co/hugging-quants/Meta-Llama-3.1-70B-Instruct-AWQ-INT4}\\
    NLLB-200-distilled-600M & \url{https://huggingface.co/facebook/nllb-200-distilled-600M}\\
    \midrule
    \multicolumn{2}{c}{\textit{Other resources}} \\
    \midrule
    MetricX23-XXL & \url{https://huggingface.co/google/metricx-23-xxl-v2p0} \\
    XCOMET-XXL & \url{https://huggingface.co/Unbabel/XCOMET-XXL} \\
    \texttt{wmt23-cometkiwi-da-xxl} & \url{https://huggingface.co/Unbabel/wmt23-cometkiwi-da-xxl} \\
    FastText & \url{https://huggingface.co/facebook/fasttext-language-identification} \\
    BM25s & \url{https://github.com/xhluca/bm25s} \\
    \bottomrule
    \end{tabular}
    }
    \label{tab:url}
\end{table*}

\subsection{Prompts} \label{appendix:prompts}

\subsubsection{Translation prompt}
Zero-shot
\begin{lstlisting}[frame=single,breaklines=true,basicstyle=\small\ttfamily]
Please write a high-quality Amharic translation of the following English sentence

"We now have 4-month-old mice that are non-diabetic that used to be diabetic," he added.

Please provide only the translation, nothing more.    
\end{lstlisting}
Few-shot
\begin{lstlisting}[frame=single,breaklines=true,basicstyle=\small\ttfamily]
Given the following sentence-translation pairs written by a professional translator:

<Demonstrations>
1. English sentence
"If it becomes commercial, we should have it. That is, there's no in-principle objection to nuclear energy" Mr Costello said.
Amharic translation
<>

2. English sentence
The governor also stated, "Today, we learned that some school aged children have been identified as having had contact with the patient."
Amharic translation
<>

3. English sentence
The commissioner said, "We haven't yet agreed on rules of origin and tariff con[c]essions, but the framework we have is enough to start trading on July 1, 2020".
Amharic translation
<>

4. English sentence
Permits are limited to protect the canyon, and become available on the 1st day of the month, four months prior to the start month.
Amharic translation
<>

5. English sentence
We have a year-long financial crisis, which has had its most acute moment in the past two months, and I think now the financial markets are beginning to recover."
Amharic translation
<>
</Demonstrations>

Please write a high-quality Amharic translation of the following English sentence

"We now have 4-month-old mice that are non-diabetic that used to be diabetic," he added.

Please make sure to consider the above information and provide only the translation, nothing more.
\end{lstlisting}

\subsubsection{Divide prompt}
Vanilla
\begin{lstlisting}[frame=single,breaklines=true,basicstyle=\small\ttfamily]
We would like to derive a list of short sentences from long and convoluted sentences. For each long sentence, you will use punctuation (e.g., comma, semicolon, etc.), coordinating conjunctions (e.g., for, and, etc.), and subordinating conjunctions (e.g., although, because) to divide the sentence into multiple clauses, which you will then use to write simpler sentences. Ensure that each of the short sentences reflects a part of the larger sentence. Here are some examples.

###

Sentence
The Boolean satisfiability problem is a well-researched problem with many exemplar solvers available; it is very fast, as package solving complexity is very low compared to other areas where SAT solvers are used. 

Propositions
    1. The Boolean satisfiability problem is a well-researched problem. 
    2. It has many exemplar solvers are available.
    3. It is very fast.
    4. The package solving complexity is very low. 
    5. This is compared to other areas where SAT solvers are used.

###

Sentence
Dore was offered several one-off shows in night clubs, and her best album was rereleased in 2001. 

Propositions
    1. Dore was offered several one-off shows in night clubs.
    2. Her best album was rereleased in 2001.

###

Sentence
Jim briefly transfers to the Stamford branch after Pam confirmed her commitment to Roy, before corporate is forced to merge the Stamford branch and staff into the Scranton branch.

Propositions
    1. Jim briefly transfers to the Stamford branch.
    2. Pam confirmed her commitment to Roy.
    3. Corporate is forced to merge the Stamford branch and staff.
    4. The merge is into the Scranton branch.

###

Sentence
But Jack could not get back to his own time, because one of the drug vials had broke, and there was only enough left in one of the vials to stop Whistler.

Propositions
    1. But Jack could not get back to his own time.
    2. One of the drug vials had broke.
    3. There was only enough left in one of the vials.
    4. This was to stop Whistler.

###

Sentence
However, his nonconformist background came to the fore again when he became friendly with William Durning around 1817, having rented a cottage from another member of the Durning family, and on 1 September 1820 he married William's daughter, Emma.

Propositions
    1. However, his nonconformist background came to the fore again.
    2. He became friendly with William Durning around 1817.
    3. He rented a cottage from another member of the Durning family.
    4. He married William's daughter.
    5. The marriage was on 1 September 1820.

###

Sentence
Mallzee was founded in December 2012 by Cally Russell and is based in Edinburgh.

Propositions
    1. Mallzee was founded in December 2012 by Cally Russell.
    2. It is based in Edinburgh.

###

Sentence
He was educated at William Ellis School before being accepted into University College London to study botany and zoology, after graduating he went to the College of the Pharmaceutical Society and studied pharmacy, graduating in 1935. 

Propositions
    1. He was educated at William Ellis School.
    2. This was before being accepted into University College London.
    3. This was to study botany and zoology. 
    4. After graduating he went to the College of the Pharmaceutical Society.
    5. He studied pharmacy.
    6. He graduated in 1935.

###

Sentence
Out of 3 other surrounding neighborhoods, Mattapan saw a population decrease but has the highest proportion of Black/African American residents in the city, but the number of blacks actually dropped over the last decade.

Propositions
    1. Out of 3 other surrounding neighborhoods.
    2. Mattapan saw a population decrease.
    3. It has the highest proportion of Black/African American residents in the city.
    4. The number of blacks actually dropped over the last decade.

###

Sentence
Nerepis is situated on the Nerepis River and is located east of the town of Grand Bay-Westfield in the Saint John, the nearest city, which is about twenty-five minutes away. 

Propositions
    1. Nerepis is situated on the Nerepis River.
    2. It is located east of the town of Grand Bay-Westfield.
    3. Grand Bay-Westfield is in the Saint John.
    4. Saint John is the nearest city.
    5. It is about twenty-five minutes from Nerepis.

###

Sentence
In 1961, when Muskee was 20 years old, his mother died, and a year later his grandmother died.

Propositions
    1. In 1961, when Muskee was 20 years old.
    2. His mother died.
    3. A year later, his grandmother died.

###

Sentence
{}
\end{lstlisting}


Paraphrase
\begin{lstlisting}[frame=single,breaklines=true,basicstyle=\small\ttfamily]
We would like to propose a list of paraphrases of sentences. For each sentence, you will provide four paraphrases that have the same meaning as the original sentence and mostly use the same words as well.
Ensure that each of the four paraphrases is a correct sentence and does not change the meaning of the original sentence.
Here are some examples.

###

Sentence
The Boolean satisfiability problem is a well-researched problem with many exemplar solvers available; it is very fast, as package solving complexity is very low compared to other areas where SAT solvers are used. 

Propositions
    1. The Boolean satisfiability problem is a widely studied topic, with numerous exemplar solvers available; it is efficient, as solving package complexity is significantly lower than in other domains using SAT solvers.
    2. Boolean satisfiability, a well-researched problem, boasts many exemplar solvers, and its speed is notable due to the low complexity of package solving compared to other SAT applications.
    3. The problem of Boolean satisfiability has been extensively researched, leading to the development of many exemplar solvers; package solving in this context is fast, given its comparatively low complexity in contrast to other SAT solver uses.
    4. With numerous exemplar solvers available, the Boolean satisfiability problem is well-researched and demonstrates remarkable speed, as the complexity of package solving is much lower than in other SAT solver applications.

###

Sentence
Dore was offered several one-off shows in night clubs, and her best album was rereleased in 2001. 

Propositions
    1. Dore's best album was rereleased in 2001, and she was offered several one-off shows in night clubs.
    2. In 2001, Dore's best album was rereleased, and she received offers for several one-off performances in night clubs.
    3. Several one-off shows in night clubs were offered to Dore, and her best album saw a rerelease in 2001.
    4. Dore was given opportunities for one-off performances in night clubs, and her best album was rereleased during 2001.

###

Sentence
Jim briefly transfers to the Stamford branch after Pam confirmed her commitment to Roy, before corporate is forced to merge the Stamford branch and staff into the Scranton branch.

Propositions
    1. After Pam confirmed her commitment to Roy, Jim briefly transfers to the Stamford branch, only for corporate to merge Stamford staff into the Scranton branch.
    2. Jim transfers briefly to the Stamford branch after Pam confirms her commitment to Roy, but corporate later merges the Stamford staff into the Scranton branch.
    3. Pam's confirmation of her commitment to Roy leads Jim to briefly transfer to the Stamford branch, which is later merged into the Scranton branch by corporate.
    4. Before corporate merges the Stamford branch and its staff into the Scranton branch, Jim briefly transfers there after Pam confirms her commitment to Roy.

###

Sentence
But Jack could not get back to his own time, because one of the drug vials had broke, and there was only enough left in one of the vials to stop Whistler.

Propositions
    1. Jack could not return to his own time because one of the drug vials had broken, leaving only enough in one vial to stop Whistler.
    2. Since one of the drug vials had broken, Jack was unable to get back to his own time, with just enough remaining in a single vial to stop Whistler.
    3. Because one of the vials of the drug had broken, Jack could not make it back to his own time, as only one vial had enough left to stop Whistler.
    4. One of the drug vials had broken, leaving Jack unable to return to his own time, with only enough left in one vial to stop Whistler.

###

Sentence
{}
\end{lstlisting}

\subsubsection{Merge prompt}
Vanilla
\begin{lstlisting}[frame=single, breaklines=true, basicstyle=\small\ttfamily]
Given the following sentence-translation pairs written by a professional translator:

<Demonstrations>
1. English sentence
The mice used to be diabetic.
Amharic translation
<>

2. English sentence
They now have 4-month-old mice.
Amharic translation
<>

3. English sentence
The mice are non-diabetic.
Amharic translation
<>
</Demonstrations>

Please write a high-quality Amharic translation of the following English sentence

"We now have 4-month-old mice that are non-diabetic that used to be diabetic," he added.

Please make sure to consider the above information and provide only the translation, nothing more
\end{lstlisting}

\subsection{About the decomposition step} \label{appendix:structural}
For structural decomposition, we use the dependency tree to recursively split the sentence. First, we identify the root of the sentence and divide it into two parts: the left part (including the root) and the right part (excluding the root). Both parts are added to a stack, ensuring that the longer segment (in terms of word count) is processed first. This process continues until all subparts contain no more than four words.

\subsection{About the existing strategies}
\paragraph{MAPS}

MAPS~\citep{he2024exploring} is an ensembling strategy where an LLM generates three different translations of a given sentence by analyzing 3 aspects. The available implementation only supports HRLs (English, Chinese, French, German, Japanese). In order to extend it to more languages, we translated their list of keywords using \texttt{NLLB-200-distilled-600M}. We found that their trigger sentences came from FLORES-200, so we used their equivalents in other FLORES languages. We also selected demonstrations from FLORES and ensured they were related to the trigger sentences.

\paragraph{SBYS}

SBYS~\citep{briakou-EtAl:2024:WMT} uses a multi-turn conversation in order to drive an LLM to output a better translation. The authors did not provide an open-source implementation of their method so we had to build it from scratch with the help of the prompts provided in the paper.

\paragraph{TEaR}
We re-implemented TEaR~\citep{feng2024tearimprovingllmbasedmachine} using the prompts provided in their paper to make it compatible with our benchmarks. Since CompTra uses BM25 as its retriever, we used it for TEaR as well. We used Few-shot Translate, Few-shot Estimate and $\mathcal{T}_{refine} - \beta$ as the Refine prompt.

\paragraph{Self-Refine}
Following \citet{chen2024iterativetranslationrefinementlarge}, we used their \textit{Refine} prompt which simply ask an LLM to refine a translation given the source sentence. They reported that it usually works best, reaching its peak performance after one iteration of refinement.

\paragraph{CoT}
For zero-shot CoT, we adopted the traditional \textit{Let's think step by step}~\citep{NEURIPS2022_8bb0d291} trigger sentence with the two-step procedure: Reasoning extraction and answer extraction. In few-shot CoT, we did not provide examples of step-by-step reasoning in the demonstrations, instead we just appended \textit{Let's think step by step} at the end of the prompt (after the few-shot demonstrations) and did the two-step procedure as in zero-shot CoT.

\section{Qualitative Comparison between the outputs of models} \label{appendix:comparison}

\begin{table*}[!ht]
    \centering\small
    \caption{\texttt{LLaMA 3.1 70B It}'s generations in Samoan.}
    \label{tab:qualitative}
    \resizebox{\linewidth}{!}{
    \begin{tabular}{p{3cm} p{12cm}}
    \toprule
    Source sentence & His work is of such recognized quality and detail that he is one of the very few "household names" among philatelists. Some specialize in collecting his work alone. \\
    \midrule
    Zero-shot & O le galuega a ia o le aofaiga ma le faʻamalieina o le faʻamaonia, o lea e faʻaigoa ai o ia i totonu o le laupesi o tagata faʻatau tupe. E toʻaitiiti tagata e faʻaaluina o latou galuega i le faʻatauina o ana galuega e tasi. \\
    \midrule
    Zero-shot + CoT & O le galuega a ia e fa'amaonia lelei ma le fa'apalepale o lea e tasi o le vae to'atele o tagata e fa'amaonia i totonu o tagata fa'afofoga pepa. E iai ni tagata e fa'afofoga i le galuega a ia e tasi. \\
    \midrule
    Zero-shot + Refine & O le galuega a ia o le aofaiga ma le faʻamalieina o le faʻamaonia, o lea e faʻaigoa ai o ia i totonu o le laupesi o tagata faʻatau tupe. E toʻaitiiti tagata e faʻaaluina o latou galuega i le faʻatauina o ana galuega e tasi. \\
    \midrule
    SBYS & O le galuega a ia o le aofaiga ma le faʻamalieina o le tulaga e faʻamaonia, o le mea lea e faʻaigoa ai o ia i totonu o le au faʻatau tupe, e leʻi tele ni tagata e faʻatau tupe e faʻaalu i le galuega a ia. \\
    \midrule
    MAPS & O lana galuega e faʻamaonia lelei ma le faʻamalieina o le aiga o le igoa o le aiga i totonu o tagata faʻatau pese. E toʻaitiiti tagata e faʻapipiʻi i le faʻatauina o lana galuega e tasi. \\
    \midrule
    TEaR & O lana galuega e faʻamaonia lelei ma le faʻamalieina, o le mea lea e faʻaalu ai o ia i totonu o le vae toʻaitiiti o "igoa o le aiga" i totonu o tagata faʻatau tupe. E toʻaitiiti e faʻapipiʻi i le faʻatauina o lana galuega e aunoa. \\
    \midrule
    5-shot BM25 & O lana galuega e faʻamaonia lelei ma le faʻamalieina o le aofaʻi, o le mea lea e faʻaigoa ai o ia o se tasi o le vae toʻaitiiti o "igoa o fale" i totonu o tagata faʻatau tupe. E toʻaitiiti e faʻaoga i le faʻatauina o lana galuega e tasi. \\
    \midrule
    5-shot BM25 + CoT & Lana galuega e fa'amaonia ma le fa'afine, o le mea lea e fa'apea o ia o se tasi o le vae to'aitiiti o igoa o le aiga i totonu o tagata fa'afoe. E iai nisi e fa'aaluina i le fa'atauina o lana galuega e tasi. \\
    \midrule
    5-shot BM25 + Refine & O lana galuega e faʻamaonia lelei ma le faʻamalieina o le aofaʻi, o le mea lea e faʻaigoa ai o ia o se tasi o tagata e sili ona lauiloa i totonu o tagata faʻatau tupe. E toʻaitiiti e faʻaoga i le faʻatauina o lana galuega e tasi. \\
    \midrule
    CompTra &  O lana galuega e fa’amaonia lelei ma le fa’afouina, o se tasi o ni vaaiga lauiloa i totonu o tagata fa’aalu i tupe. E i ai nisi e sili ona fa’aauau i le fa’asoa o lana galuega e tasi. \\
    \midrule
    Reference & O lana galuega sa lauiloa i le tulaga lelei ma sa lauiloa lona igoa i le lisi o e faia faailoga o tusi (stamps). E i ai tagata faapitoa i le aoina mai ana galuega. \\
    \bottomrule
    \end{tabular}
    }
\end{table*}

\end{document}